\documentclass[journal]{IEEEtran}
\usepackage[T1]{fontenc}
\usepackage[cmintegrals]{newtxmath}
\usepackage{graphicx}
\usepackage{color} 
\usepackage{epsf}
\usepackage{amsmath, amsfonts}
\usepackage{latexsym}
\usepackage{subfigure}
\usepackage{multirow}
\usepackage{multicol}
\usepackage{cite}
\usepackage{float}
\usepackage{algorithm}
\usepackage{algpseudocode}
\usepackage{lipsum}
\usepackage{cleveref}

\newtheorem{lemma}{Lemma}
\newtheorem{assumption}{Assumption}
\newtheorem{corollary}{Corollary}
\newtheorem{theorem}{Theorem}
\newtheorem{proposition}{Proposition}
\newtheorem{remark}{Remark}

\newcommand{\titerra}{T^{\text{iter, RA}}}
\newcommand{\taucomm}{\tau^{\text{comm}}}
\newcommand{\taucomp}{\tau^{\text{comp}}}
\newcommand{\psuc}{p^{\text{suc}}}
\newcommand{\ptr}{p^{\text{tr}}}
\newcommand{\tcomm}{T^{\text{comm}}}
\newcommand{\navail}{N^{\text{avail}}}

\newcommand{\argmin}{\operatornamewithlimits{argmin}}

\ifCLASSINFOpdf
\else
\fi
\usepackage{amsmath}
\usepackage[cmintegrals]{newtxmath}
\begin{document}
	
	\title{Optimal Batch Allocation for Wireless Federated Learning}
	
	\author{Jaeyoung~Song,~\IEEEmembership{Member,~IEEE} and Sang-Woon Jeon,~\IEEEmembership{Senior Member,~IEEE}
		\thanks{J. Song is with the Department of Electronics Engineering, Pusan National University, Pusan, 46241, Korea (e-mail : jsong@pnu.edu)} \\
		\thanks{S.-W. Jeon is with the Department of Electrical and Electronic
			Engineering, Hanyang University, Ansan 15588, South Korea (e-mail:
			sangwoonjeon@hanyang.ac.kr ).}}
	
	
	\maketitle
	
	\begin{abstract}
		Federated learning aims to construct a global model that fits the dataset distributed across local devices without direct access to private data, 
		leveraging communication between a server and the local devices. In the context of a practical communication scheme, we study the completion 
		time required to achieve a target performance. Specifically, we analyze the number of iterations required for federated learning to reach a specific 
		optimality gap from a minimum global loss. Subsequently, we characterize the time required for each iteration under two fundamental multiple 
		access schemes: time-division multiple access (TDMA) and random access (RA). We propose a step-wise batch allocation, demonstrated to be 
		optimal for TDMA-based federated learning systems. Additionally, we show that the non-zero batch gap between devices provided by the 
		proposed step-wise batch allocation significantly reduces the completion time for RA-based learning systems. Numerical evaluations validate 
		these analytical results through real-data experiments, highlighting the remarkable potential for substantial completion time reduction.
	\end{abstract}
	
	\begin{IEEEkeywords}
		Batch allocation, federated learning, multiple access, wireless distributed learning.
	\end{IEEEkeywords}
	
	\IEEEpeerreviewmaketitle
	\section{Introduction}
	The rapid advancement of Internet of Things (IoT) technology has led to an explosive increase in the number of IoT devices. It is estimated that the 
	number of IoT connections will surpass 34 billion by 2023, according to Ericsson's recent report \cite{Ericsson_report2023}. This exponential growth 
	in IoT devices has resulted in the generation of vast volumes of data. Simultaneously, the dramatic evolution of machine learning techniques, 
	particularly in the realm of artificial intelligence (AI), has opened new avenues for harnessing the wealth of data generated by these devices.
	
	However, due to the vast scale of IoT datasets, constructing a single centralized machine for storage and processing becomes impractical. 
	Furthermore, numerous types of data generated by IoT, such as financial and health information, are personal and highly sensitive. This naturally 
	results in data owners being hesitant to grant access to centralized entities.
	To address such challenges of handling large-scale data while preserving data privacy, federated learning has emerged as a promising solution 
	\cite{McMahan_MLR2017}. In a federated learning system, a central server coordinates the learning process, while local devices compute models, 
	i.e., artificial neural network models, using their own locally stored data. Importantly, no data is exchanged between the server and the devices. 
	Instead, the local models computed by the devices are transmitted to the server. Upon receiving these local models, the server aggregates them to 
	produce a global model, which is subsequently sent back to the devices for the next iteration. This iterative process continues, ultimately resulting in 
	the training of the global model.
	
	One of the key advantages of federated learning is that it does not require direct access to local data. The server constructs the global model by 
	aggregating local models, making it well-suited for applications involving sensitive or private data. Furthermore, since the entire dataset is distributed 
	across numerous devices, the computational load on each device is substantially lower than that in centralized learning. This distributed nature of 
	federated learning also enables the exploitation of parallel computing techniques \cite{Kairouz_NOW2021}.
	
	However, a limitation of federated learning is that no single entity can access the entire dataset. Consequently, updating the global model in each 
	iteration may be less efficient compared to centralized counterparts, often requiring more iterations to achieve the desired performance. Additionally, 
	communication is essential in each iteration, as local models must be transmitted from the devices to the server, and the updated global model is 
	sent from the server to the devices. This communication overhead can become a bottleneck in federated learning \cite{Zhang_IOTM2022}.
	
	In response to these challenges, researchers have focused on developing communication-efficient federated learning techniques. Notable 
	approaches include Federated Averaging (FedAvg), which aggregates local models from a subset of devices to reduce communication overhead 
	\cite{McMahan_MLR2017}. Moreover, in  \cite{Konecny_NIPS2016}, the authors have proposed an updates scheme that compresses parameters of 
	local models to reduce communication cost.
	In addition, strategies involving periodic transmission of local models have been proposed in \cite{Stich_Arxiv2018, Yu_AAAI2019} to decrease 
	communication frequency. Some methods have even leveraged primal-dual optimization techniques to further reduce communication requirements 
	\cite{Lan_MP2020}.
	
	Another approach seeks to minimize the number of communications by introducing event-triggered aggregation, where local models are sent to the 
	server only when the difference between the most recently sent model and the current local model exceeds a predefined threshold 
	\cite{Liu_TAC2019}. This approach aligns with the development of decentralized systems that address optimization problems in a distributed manner 
	\cite{Gao_TNSE2022}.

	In addition to that, another line of research focuses on resource optimization for enhancing bandwidth and energy efficiency. Energy-efficient 
	strategies have considered the amount of energy used in both transmission and computation, leading to solutions that minimize energy consumption 
	for federated learning \cite{Yang_TWC2020, Cao_JSAC2022}. Bandwidth allocation schemes that select users with a probability distribution have 
	also been explored in \cite{Chen_TWCC2020}. Further work in \cite{Wan_JSAC2021}  has aimed to jointly optimize power and bandwidth allocation 
	to minimize energy and time requirements. Furthermore, the impact of the number of participating devices on communication costs and computing 
	loads has been studied in \cite{Song_JSAC2020}, with an exploration of the optimal number of devices for federated learning systems.
	
	To enhance the efficiency of model training, adaptive batch sizes to accelerate convergence behavior and reduce the number of required iterations 
	have been studied in \cite{Devra_Arxiv2017, Zhang_TCOMP2022}. In particular, techniques such as deep reinforcement learning have been 
	employed to determine batch sizes, aligning them with the capabilities of individual devices. Furthermore, optimizing the exponential factor 
	governing batch size increases has been explored to minimize the required time of federated learning in \cite{Shi_TWC2022}.
	
	It is worth mentioning that existing literature often assumes uniform batch sizes across all participating devices, resulting in the simultaneous 
	transmission of their locally updated models. This can lead to excessive interference and congestion for local model transmission at each device, 
	particularly when dealing with a large number of devices. In light of these challenges, our research is centered on heterogeneous batch allocation, 
	where different batch sizes are assigned to individual devices. This heterogeneous allocation naturally introduces variability in the time duration 
	required for updating local models on different devices, thereby reducing the effective number of devices accessing the wireless channel and 
	mitigating communication overhead. To the best of our knowledge, strategies that allocate different batch sizes to each of devices have not been 
	thoroughly investigated. Additionally, we consider two fundamental multiple access schemes: time-division multiple access (TDMA) and random 
	access (RA) for transmitting local models to the server. 
	For clarity, our contributions are summarized as follows.
	\begin{itemize}
		\item The completion time of federated learning is characterized, representing the time required for achieving a specified optimality gap from the 
		optimal model. To accomplish this goal, the number of iterations necessary to guarantee the desired optimality gap is firstly determined. 
		Subsequently, the expression for the time duration of each iteration is derived, considering two distinct multiple access schemes.
		\item A novel step-wise batch allocation algorithm is proposed that is specifically designed to minimize the iteration time within federated learning 
		scenarios sharing the same wireless resource among multiple devices. It is proved that the proposed batch allocation is optimal when TDMA is 
		utilized.
		\item When RA is applied, optimal batch allocation strategies are identified for the two-device case and necessary and sufficient conditions for 
		optimal batch allocation in the three-device case are established. 
		For scenarios involving a general number of devices, the step-wise batch allocation is shown to significantly reduce the completion time of 
		federated learning compared to the conventional equal-sized batch allocation.
		\item The performance of the proposed batch allocation algorithms is rigorously evaluated through a series of extensive experiments. The 
		significant impact of appropriately spacing batch sizes between devices is underscored, leading to a substantial reduction in completion time for 
		both TDMA-based and RA-based federated learning systems.
	\end{itemize}
	
	The rest of this paper is organized as follows. In Section~\ref{sec:system_model}, we formally formulate the completion time minimization for 
	federated learning systems. In Section~\ref{sec:required iterations}, we characterize the completion time of federated learning achieving a specified 
	optimality gap from the optimal model. Then, we analyze the iteration time in Section~\ref{sec:iter_time} and optimal batch location in 
	Section~\ref{sec:opt_batch} for both TDMA and RA protocols. Moreover, we verify our analysis via numerical results in Section~\ref{sec:exp}. Finally, 
	Section~\ref{sec:conclusion} concludes the paper.

	\section{Problem Formulation}\label{sec:system_model}
	\subsection{Federated Learning System}
	\begin{figure}[h!]
		\centering
		\includegraphics[scale=0.3]{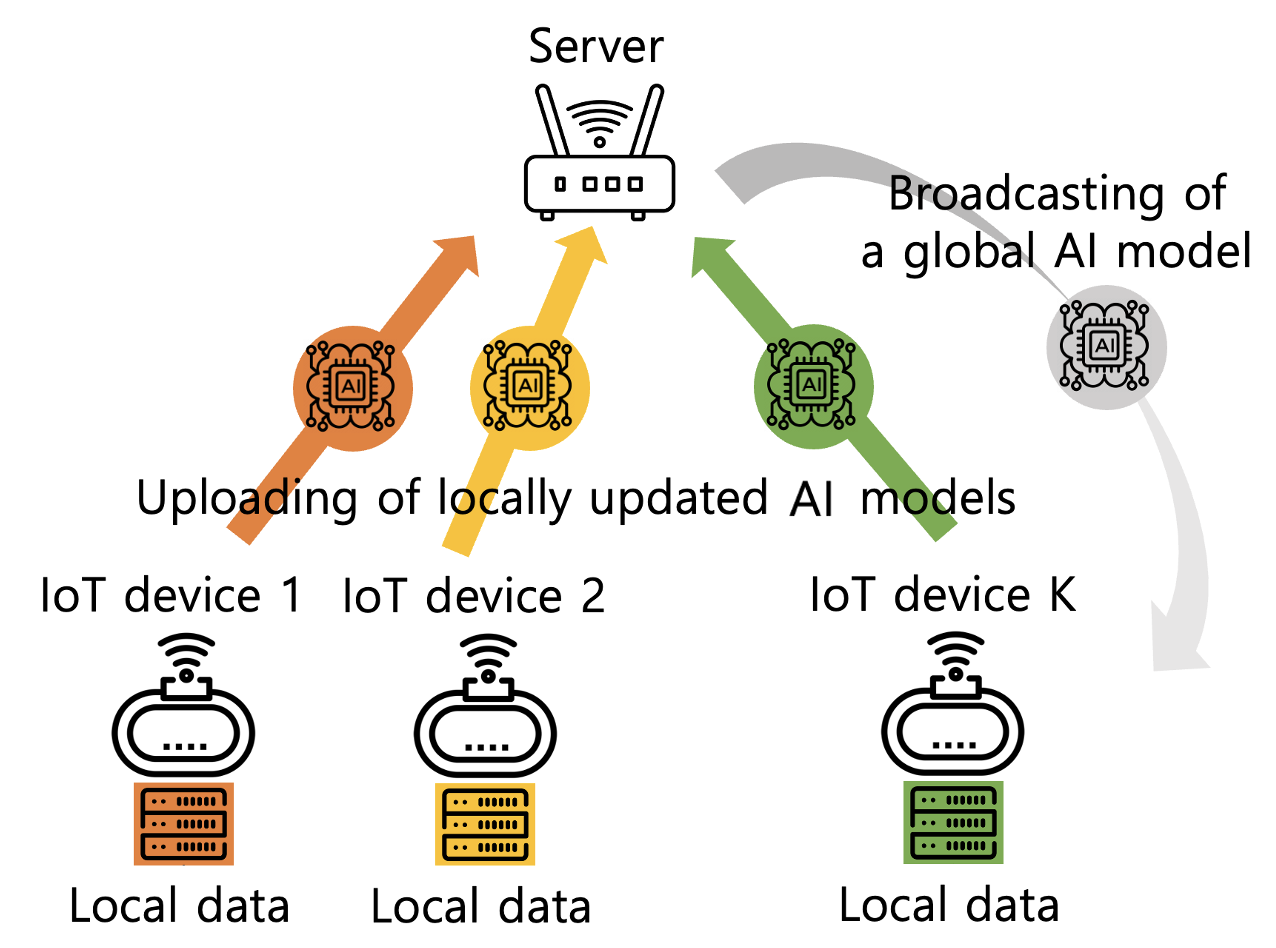}
		\caption{Federated learning system with a server and  $N$ IoT devices.}\label{fig:system_model}
	\end{figure}
	We consider a federated learning system depicted in Fig. \ref{fig:system_model} consisting of a single server and $N$ IoT devices. Each device $n$ 
	has its own local data set $\mathcal{D}_n$, where $n\in[1:N]$. Then, the entire data set in this system is represented as $\mathcal{D} = 
	\bigcup_{n=1}^N \mathcal{D}_n$. The aim of the federated learning system is to find an optimal global model which fits to the entire data set without 
	directly accessing to local data. Specifically, let $\overline{\mathbf{w}}$ denote a global model and its global loss function is defined as
	\begin{align}
		f(\overline{\mathbf{w}}) = \frac{1} {|\mathcal{D}|} \sum_{n=1}^N |\mathcal{D}_n| f_n(\overline{\mathbf{w}}; \mathcal{D}_n), \label{def:global_loss}
	\end{align} where $f_n(\overline{\mathbf{w}}; \mathcal{D}_n)$ is the local loss function for device $n$ measured by the local data set 
	$\mathcal{D}_n$. In other words,
	\begin{align}
		f_n(\overline{\mathbf{w}} ; \mathcal{D}_n) = \frac{1}{|\mathcal{D}_n|} \sum_{\zeta \in \mathcal{D}_n} l(\overline{\mathbf{w}} ; \zeta),
	\end{align}  where $l(\overline{\mathbf{w}}; \zeta)$ is the loss function for each data sample $\zeta$.
	Hence, an optimal model can be expressed as 
	\begin{align} \label{eq:w_star}
		\overline{\mathbf{w}}^* = \argmin_{\overline{\mathbf{w}}} f(\overline{\mathbf{w}})
	\end{align}
	and the corresponding minimum global loss is given by $f^*=f(\overline{\mathbf{w}}^*)$.
	
	To protect local data stored at the devices, the server is not allowed to directly access to local data stored at the devices. Hence, each device is 
	required to compute and update local model based on its own local data set. After updating, the updated local model will be 
	delivered to the server. 
	The server then receives the updated local models from the devices and aggregates them to produce a global model for the next iteration. By 
	iterating such local update at each device and aggregation at the server, the federated learning system can obtain an optimal global model without 
	accessing local data stored in $N$ devices.
	
	Due to the large scale of data stored on devices, we use stochastic gradient descent for updating models. Stochastic gradient descent involves 
	computing gradients with randomly selected data samples. The subset of data samples chosen for computing gradients is called a batch. Thus, if 
	device $n$ utilizes batch $\mathcal{B}_n^k\subset \mathcal{D}_n$ of size $B_n^k$ at the $k$th iteration, the stochastic gradient $\mathbf{g}_n^k$ 
	is represented as
	\begin{align} \label{eq:sg}
		\mathbf{g}_n^k &= \nabla f_n ( \overline{\mathbf{w}}^k ; \mathcal{B}_n^k)= \frac{1}{B_n^k} \sum_{\zeta \in \mathcal{B}_n^k} \nabla 
		l(\overline{\mathbf{w}}^k ; \zeta),
	\end{align} where $\overline{\mathbf{w}}^k$ denotes the global model at the $k$th iteration. Without loss of generality, we assume that the batch 
	size is assigned in ascending order with respect to the device index $n$, i.e., $B^k_n \leq B^k_{n'}$ if $n\leq n'$.
	We assume that the stochastic gradient is an unbiased estimator of the true gradient. In other words,
	\begin{align}
		\mathbb{E}_{\zeta} [ \mathbf{g}^k_n ] = \nabla f( \overline{\mathbf{w}}^k) \label{eq:unbiased}
	\end{align}
	is satisfied, where the expectation takes with respect to $\zeta$ which is a chosen data sample at the $k$th iteration.
	By using the stochastic gradient $\mathbf{g}_n^k$ in \eqref{eq:sg}, device $n$ updates the global model $\overline{\mathbf{w}}^k$ at the $k$th 
	iteration as 
	\begin{align}
		\mathbf{w}_n^{k+1} = \overline{\mathbf{w}}^k - \eta^k \mathbf{g}_n^k, \label{eq:sgd}
	\end{align}
	where $\eta^k>0$ is the step size at the $k$th iteration. 
	
	Then, after receiving $\mathbf{w}^{k+1}_n$ for all $n\in[1:N]$ at the server, the global model $\overline{\mathbf{w}}^{k+1}$ for the $(k+1)$th 
	iteration is constructed as
	\begin{align}
		\overline{\mathbf{w}}^{k+1} = \sum_{n=1}^N \frac{B^k_n}{B} \mathbf{w}_n^{k+1}, \label{eq:aggregation}
	\end{align} where $B = \sum_{n=1}^N B^k_n$.
	Note that $\mathbf{w}^{k+1}_n$ is an updated local model at device $n$ and is different from a global model $\overline{\mathbf{w}}^{k+1}$, which 
	is the output of aggregation at the server.
	\begin{remark}
		The main focus of this paper is about batch allocation and the corresponding transmission control between $N$ devices when the same number of 
		data samples for batches is assumed at each iteration. For notational simplicity, therefore, denote $B = \sum_{n=1}^N B^k_n$ by ignoring the 
		index $k$. \hfill$\lozenge$
	\end{remark}
	
	Through the repetition of local updates on devices and global aggregation at the server, an optimal global model can be obtained. In this paper, our 
	focus is on the completion time required to obtain an updated global model at the server, approaching the optimal state within a certain gap. To 
	measure the completion time, we adopt a time-slotted model where each time slot is defined as a specific duration. Consequently, by counting the 
	number of time slots needed to establish the optimal model, we can measure the completion time for federated learning. 
	
	\subsection{Minimization of Completion Time} \label{subsec:min_completion_time}
	
	In this paper, the completion time of the federated learning system is defined as the time duration for constructing 
	$\overline{\mathbf{w}}^k$ that satisfies
	\begin{align}
		\mathbb{E} \left[ f(\overline{\mathbf{w}}^k)  \right] - f^* \leq \epsilon,  \label{ineq:opt_gap}
	\end{align} where $\epsilon>0$ is a target optimality gap.
	Provided that a model at the $K(\epsilon)$th iteration satisfies \eqref{ineq:opt_gap}, the completion time $T$ is represented as
	\begin{align}
		T = \sum_{k=1}^{K(\epsilon)} T^{\text{iter}}_k, \label{eq:compl_t_tdma}
	\end{align} where $T^{\text{iter}}_k$ is the time duration for updating at the $k$th iteration.
	
	In federated learning, each iteration consists of four steps: 1) distribution of a global model from the sever 
	to each 
	device; 
	2) 	local updating at each device based on its own data; 3) transmission of an updated model from each device to the server; 4) aggregation of 
	updated models at the server. For model distribution, the server can broadcast an aggregated model to all devices using a downlink shared channel. 
	Additionally, the computing capability of the server is generally much better than that of IoT devices. Therefore, we focus on steps 2 and 3 to 
	evaluate $T^{\text{iter}}_k$. 
	
	Let $T^{\text{iter}}_{k,n}$ represent the required time duration for device $n$ at the $k$th iteration. 
	Then. $T^{\text{iter}}_k=\max_{n\in[1:N]} T^{\text{iter}}_{k,n}$.
	The aim of this paper is to minimize the expected completion time that is represented as
	\begin{align} \label{eq:min_expected_time}
		&\min_{\{B_n^k\}_{n\in[1:N],k\in[1:K(\epsilon)]}}\mathbb{E} \left[T\right] \nonumber\\
		&= \min_{\{B_n^k\}_{n\in[1:N],k\in[1:K(\epsilon)]}}\left\{\sum_{k=1}^{K(\epsilon)} \mathbb{E} 
		\left[\max_{n\in[1:N]}\left\{T^{\text{iter}}_{k,n}\right\}\right]\right\},
	\end{align}
	where the expectation accounts for the randomness of batch constructions and communications depending on transmission 
	protocols.
	
	\subsubsection{Computing time for local updates}
	
	We assume that the computing capability of $N$ devices are the same such that each device can compute gradient of $\rho$ data samples per time 
	slot.  
	Recall that the batch size of device $n$ at the $k$th iteration is given by $B_n^k$. Hence, if we represent the computing time of device $n$ at the 
	$k$th iteration as $\taucomp_n$, we have
	\begin{align}
		\taucomp_{n,k} = \left\lceil \frac{B^k_n}{\rho} \right\rceil, \label{eq:t_comp}
	\end{align} 
	where $\left\lceil \cdot \right\rceil$ is the ceiling function.

	\subsubsection{Communication time for model transmission} \label{subsubsec:comm_time}
	Description of transmission protocols of $N$ nodes for updated model delivery to the server is more complicated than the computing time 
	description.
	For simplicity, we assume that each slot duration is normalized such that a single model can be transmitted from a device to the server. However, as 
	multiple devices participate in federated learning, a multiple access protocol  between devices is required to send their updated models. In this paper, 
	we consider two fundamental multiple access schemes: TDMA and RA.
	
	In the case of TDMA, a central scheduler can allocate distinct time slots for each of the devices, thereby avoiding interference or collisions that might 
	occur due to simultaneous transmissions from multiple devices. On the other hand, RA allows each device to decide 
	its transmission in a distributed manner, so that collisions can occur. Thus, each device is required to repeatedly transmit its updated model with a 
	certain 
	probability until successful transmission occurs. The detailed analysis on the communication time for both TDMA and RA will 
	be given in Section 
	\ref{sec:iter_time}. 
	
	\section{Characterization of the Number of Required Iterations} \label{sec:required iterations}
	From the definition in \eqref{ineq:opt_gap}, $K(\epsilon)$ is the expected number of iterations required for constructing a global model that can 
	achieve a minimum global loss within a constant gap of $\epsilon$. In this section, we characterize $K(\epsilon)$ when the target optimality gap 
	$\epsilon$ is given.
	To analyze the convergence behavior of the federated learning system, we introduce the following assumptions that hold for arbitrary models 
	$\mathbf{w}$ and $\mathbf{v}$.
	\begin{assumption} \label{assume:L_smooth}
		The global loss function $f$ is $L$-smooth. Specifically,
		\begin{align}
			\| \nabla f(\mathbf{w}) - \nabla f(\mathbf{v}) \| \leq L \| \mathbf{w} - \mathbf{v} \|
		\end{align}
		for any $\mathbf{w}$ and $\mathbf{v}$. \hfill$\lozenge$
	\end{assumption}
	
	From Assumption \ref{assume:L_smooth}, the following inequality holds for any $\mathbf{w}$ and $\mathbf{v}$:
	\begin{align}
		f(\mathbf{w}) \leq f(\mathbf{v}) + \left\langle \nabla f(\mathbf{v}), \mathbf{w} - \mathbf{v} \right\rangle + \frac{L}{2} \| \mathbf{w} - \mathbf{v} \|^2, 
		\label{ineq:smooth}
	\end{align} where $ \left\langle \mathbf{w} , \mathbf{v} \right\rangle$ is the inner product of $\mathbf{w}$ and $\mathbf{v}$. 
	\begin{assumption} \label{assume:M_convex}
		The global loss function $f$ is $M$-strongly convex. Specifically,
		\begin{align}
			f(\mathbf{w}) \geq f(\mathbf{v}) + \left\langle \nabla f(\mathbf{v}), \mathbf{w} - \mathbf{v} \right\rangle+ \frac{M}{2} \| \mathbf{w} - \mathbf{v} 
			\|^2 \label{ineq:str_convex} 
		\end{align} 
		for any $\mathbf{w}$ and $\mathbf{v}$. \hfill$\lozenge$
	\end{assumption}
	
	\begin{assumption}\label{sg_bound}
		For some constant $\lambda^2>0$, the mean squared norm of stochastic gradients are upper-bounded by	
		\begin{align}
			\mathbb{E} \left[ \| \nabla f_n(\mathbf{w} ; \zeta) \|^2 \right] \leq \lambda^2 
		\end{align}
		for all $n\in[1:N]$. \hfill$\lozenge$
	\end{assumption}
	
	From Assumption \ref{assume:M_convex}, the following Lemma holds.
	\begin{lemma}\label{lem:opt_gap}
		Suppose that $f( \mathbf{w})$ is $M$-strongly convex. Then, for any $\mathbf{w}$,
		\begin{align}
			2 M ( f(\mathbf{w}) -  f^*) \leq \| \nabla f(\mathbf{w}) \|^2.
		\end{align}
	\end{lemma}
	\begin{IEEEproof}
		From  Assumption \ref{assume:M_convex}, \eqref{ineq:str_convex} holds. The right-hand side of \eqref{ineq:str_convex} is minimized when 
		$\mathbf{w} = \mathbf{v} - \frac{1}{M} \nabla f(\mathbf{v})$. Thus,
		\begin{align}
			f(\mathbf{w}) &\geq f(\mathbf{v}) + \left\langle \nabla f(\mathbf{v}), \mathbf{w } - \mathbf{v} \right\rangle + \frac{M}{2} \| \mathbf{v} - \mathbf{w} 
			\|^2 \nonumber\\
			&\geq f(\mathbf{v}) - \frac{1}{M} \| \nabla f(\mathbf{v}) \|^2 + \frac{M}{2} \left\| \frac{1}{M} \nabla f (\mathbf{v}) \right\|^2 \nonumber\\
			& = f(\mathbf{v}) - \frac{1}{2 M} \left\| \nabla f(\mathbf{v} )\right\|^2 . \label{ineq:temp1}
		\end{align}
		By rewriting \eqref{ineq:temp1},
		\begin{align}
			2 M \left( f(\mathbf{v}) - f(\mathbf{w}) \right) \leq \| \nabla f(\mathbf{v})\|^2 \label{ineq:lem2_temp1}
		\end{align}
		holds for any $\mathbf{w}$. Thus, by replacing $\mathbf{w}$ with $\mathbf{w}^*$ in \eqref{eq:w_star}, 
		\begin{align}
			2M \left( f(\mathbf{v}) - f^* \right) \leq \| \nabla f(\mathbf{v})\|^2, \label{ineq:opt_grad}
		\end{align}
		which completes the proof.
	\end{IEEEproof}

	From Assumption \ref{sg_bound}, we can obtain the following lemma which bounds the second moment of stochastic gradients dependent on the 
	batch size.
	\begin{lemma}\label{lem:grad_bound}
		Recall that $\mathbf{g}_n^k$ is the stochastic gradient computed at device $n$ in the $k$th iteration with batch size $B_n^k$. From Assumption 
		\ref{sg_bound}, the second moment of $\mathbf{g}_n^k$ is upper-bounded as
		\begin{align}
			\mathbb{E} \left[ \| \mathbf{g}^k_n \|^2 \right] \leq \frac{\lambda^2}{B_n^k}
		\end{align}
		for all $n\in[1:N]$ and $k\in[1:K(\epsilon)]$.
	\end{lemma}
	\begin{IEEEproof} From \eqref{eq:sg}, we have
		\begin{align}
			\left\| \mathbf{g}_n^k \right\|^2  &=  \left\|\nabla f_n(\overline{\mathbf{w}}^k ; \mathcal{B}_n^k)\right\| ^2   \nonumber\\
			&= \left\| \nabla \left( \frac{1}{B_n^k}  \sum_{\zeta \in \mathcal{B}_n^k} f_n(\overline{\mathbf{w}}^k ; \zeta) \right) \right\|^2 \nonumber\\
			& \leq  \frac{1}{(B^k_n)^2}  \sum_{\zeta \in \mathcal{B}_n^k}  \left\| \nabla \left( f_n (\overline{\mathbf{w}}^k ; \zeta) \right) \right\|^2 .
		\end{align}
		By taking the expectation and using Assumption \ref{sg_bound},
		\begin{align}
			\mathbb{E} \left[ \| \mathbf{g}^k_n \|^2 \right] &\leq \frac{1}{(B^k_n)^2}  \sum_{\zeta \in \mathcal{B}_n^k} \mathbb{E} \left[  \left\| \nabla \left( f_n 
			(\overline{\mathbf{w}}^k ; \zeta) \right) \right\|^2 \right] \nonumber\\
			& = \frac{\lambda^2}{B_n^k},
		\end{align}
		which completes the proof.
	\end{IEEEproof}

	Now, using the given assumptions and lemmas, we can find the number of iterations required to achieve the optimality gap $\epsilon$.

	\begin{theorem}\label{thm_conv}
		Set the sequence of step sizes as 
		\begin{align}
			\eta^k = \frac{c}{\gamma + k}  \label{eq:def_eta}
		\end{align} for some constants $c > \frac{1}{M}$ $\gamma >0$, and $\eta^1 \leq \frac{1}{L}$. 
		Suppose that $\overline{\mathbf{w}}^k$ is the global model at the $k$th iteration. Then
		\begin{align}
			\mathbb{E} \left[ f(\overline{\mathbf{w}}^k) \right] - f^*  \leq \frac{\nu}{\gamma + k} \label{ineq:thm_1}
		\end{align}
		is satisfied, where  
		\begin{align}
			\nu = \max \left\lbrace \frac{L c^2 \lambda^2 N }{2B ( 2c M - 1)}, (\gamma +1 )(f(\overline{\mathbf{w}}^1) - f^*) \right\rbrace .
		\end{align}
	\end{theorem}
	\begin{IEEEproof}
		From the $L$-smoothness in \eqref{ineq:smooth},
		\begin{align}
			&	f(\overline{\mathbf{w}}^{k+1})-f(\overline{\mathbf{w}}^{k})  \nonumber\\ 
			&\leq \left\langle  \nabla f(\overline{\mathbf{w}}^k)  , \overline{\mathbf{w}}^{k+1} - \overline{\mathbf{w}}^k \right\rangle + \frac{L}{2} \left\| 
			\overline{\mathbf{w}}^{k+1} - \overline{\mathbf{w}}^k  \right\| ^2  \nonumber\\
			& = \left\langle  \nabla f(\overline{\mathbf{w}}^k)  , -\eta^k \sum_{n=1}^N \frac{B_n^k}{B} \mathbf{g}^k_n \right\rangle + \frac{L}{2} \left\| \eta^k 
			\sum_{n=1}^N \frac{B_n^k}{B} \mathbf{g}^k_n  \right\| ^2  \nonumber\\
			& = - \eta^k \sum_{n=1}^N \frac{B_n^k}{B} \left\langle  \nabla f(\overline{\mathbf{w}}^k) ,  \mathbf{g}^k_n \right\rangle + \frac{L}{2} \left\| \eta^k 
			\sum_{n=1}^N \frac{B_n^k}{B} \mathbf{g}^k_n  \right\| ^2,
		\end{align}
		where the first equality holds from \eqref{eq:sgd} and \eqref{eq:aggregation}.
		Then, by taking the expectation with respect to $\zeta = \left\lbrace \mathcal{B}^k_n | 1 \leq n \leq N \right\rbrace$ defined as 
		batches chosen at the $k$th iteration by all devices, we have
		\begin{align} \label{eq:upper_bound1}
			&\mathbb{E}_{\zeta} \left[ f(\overline{\mathbf{w}}^{k+1}) \right] - f(\overline{\mathbf{w}}^{k}) \nonumber\\
			&\leq - \eta^k \sum_{n=1}^N \frac{B_n^k}{B} \left\langle  \nabla f(\overline{\mathbf{w}}^k)  , \mathbb{E}_{\zeta} \left[  \mathbf{g}^k_n \right]  
			\right\rangle + \frac{L}{2} \mathbb{E}_{\zeta} \left[  \left\| \eta^k \sum_{n=1}^N \frac{B_n^k}{B} \mathbf{g}^k_n \right\|^2 \right]  \nonumber\\
			&\leq  - \eta^k  \left\|   \nabla f(\overline{\mathbf{w}}^k)\right\|^2 + \frac{L}{2} \mathbb{E}_{\zeta} \left[  \left\|  \eta^k \sum_{n=1}^N 
			\frac{B_n^k}{B} \mathbf{g}^k_n \right\|^2 \right] ,
		\end{align}		
		where the second inequality holds from  \eqref{eq:unbiased}. Here, the expectation is over the distribution of data selected for the $k$th iteration	
		and $\overline{\mathbf{w}}^k$ becomes deterministic at the $k$th iteration.
		
		The second term in \eqref{eq:upper_bound1} is further upper-bounded by
		\begin{align} \label{eq:upper_bound2}
			\mathbb{E}_{\zeta} \left[ \left\| \eta^k \sum_{n=1}^N \frac{B_n^k}{B} \mathbf{g}^k_n  \right\| ^2 \right] &\leq \sum_{n=1}^N \mathbb{E}_{\zeta} 
			\left[ \left\| \eta^k  \frac{B_n^k}{B} \mathbf{g}^k_n  \right\| ^2 \right]  \nonumber\\
			& = \sum_{n=1}^N (\eta^k)^2  \left(\frac{B_n^k}{B}\right)^2  \mathbb{E}_{\zeta} \left[ \left\| \mathbf{g}^k_n  \right\| ^2 \right]\nonumber\\
			& \overset{(a)}{\leq} (\eta^k)^2 \sum_{n=1}^N \frac{B_n^k}{B} \mathbb{E}_{\zeta} \left[ \left\| \mathbf{g}^k_n  \right\| ^2 \right]  \nonumber\\
			& \overset{(b)}{\leq} (\eta^k)^2 \sum_{n=1}^N \frac{B_n^k}{B} \frac{\lambda^2}{B_n^k}\nonumber\\
			& = (\eta^k \lambda)^2  \frac{N}{B},
		\end{align}
		where $(a)$ holds since $\frac{B_n^k}{B} \leq 1$ for all $n$ and  $(b)$ holds from Lemma \ref{lem:grad_bound}.
		Hence, from \eqref{eq:upper_bound1} and \eqref{eq:upper_bound2}, we have
		\begin{align} \label{eq:upper_bound3}
			&\mathbb{E}_{\zeta} \left[ f(\overline{\mathbf{w}}^{k+1}) \right] -  f(\overline{\mathbf{w}}^{k}) \nonumber\\
			&\leq   -\eta^k \| \nabla f(\overline{\mathbf{w}}^k) \|^2 + \frac{L}{2} (\eta^k \lambda)^2  \frac{N}{B} \nonumber\\
			& \leq -2 \eta^k M ( f(\overline{\mathbf{w}}^k) - f^*) + \frac{L}{2} (\eta^k  \lambda)^2 \frac{N}{B}, 
		\end{align}
		where the result in Lemma \ref{lem:opt_gap} is used.
		
		By taking the expectation for both sides of \eqref{eq:upper_bound3} over the whole sequence of samples chosen until the $k$th iteration, the 
		following inequality is obtained after some manipulations:
		\begin{align}
			&\mathbb{E}\left[ f(\overline{\mathbf{w}}^{k+1}) \right] - f^* \nonumber\\
			&\leq (1 - 2 \eta^k M ) \left(\mathbb{E} \left[ f(\overline{\mathbf{w}}^k) \right] - f^* \right) + \frac{L}{2} (\eta^k \lambda)^2 \frac{N}{B}. 
			\label{ineq:thm_1_temp}
		\end{align}
		
		Now, we are ready to prove \eqref{ineq:thm_1} by induction based on \eqref{ineq:thm_1_temp}.
		Suppose that \eqref{ineq:thm_1} holds for some $k\geq 2$. By applying \eqref{eq:def_eta} and \eqref{ineq:thm_1}, \eqref{ineq:thm_1_temp} 
		becomes
		\begin{align}
			&\mathbb{E} \left[ f(\overline{\mathbf{w}}^{k+1}) \right] - f^* \nonumber\\
			&\leq \left( 1 - 2  \frac{cM}{\gamma + k} \right) \frac{\nu}{\gamma + k} + \frac{L}{2} \left( \frac{c \lambda}{\gamma + k}\right) ^2 \frac{N}{B} 
			\nonumber\\
			& = \frac{\gamma + k - 1}{(\gamma + k)^2} \nu + \frac{1}{(\gamma + k)^2} \left( \frac{ L c^2 \lambda^2 N}{2B} - (2 c M -1)\nu \right) \nonumber\\  
			& \overset{(a)}{\leq} \frac{\gamma + k - 1}{(\gamma + k)^2} \nu \nonumber\\ 
			&\overset{(b)}{\leq} \frac{\nu}{\gamma +  k +1}, 
			\label{ineq:thm_1_temp2}
		\end{align}
		where $(a)$ holds from the definition of $\nu$ that ensures
		\begin{align}
			\frac{1}{(\gamma + k)^2} \left(  \frac{ L c^2 \lambda^2 N }{2B} - (2c M - 1 ) \nu  \right)  \leq  0
		\end{align}	
		and $(b)$ holds from $(\gamma + k)^2  = (\gamma + k + 1)(\gamma + k -1) +1$. Note that \eqref{ineq:thm_1_temp2} is the same form as in 
		\eqref{ineq:thm_1} by replacing $k$ with $k+1$.
		
		Now, let us consider the case where $k=1$. From the definition of $\nu$, it is obvious that \eqref{ineq:thm_1} holds. Therefore, \eqref{ineq:thm_1} 
		holds for all $ k \geq 1$, which completes the proof.
	\end{IEEEproof}
	
	\begin{corollary}\label{col:n_iter_req}
		Given $\epsilon$, $\mathbb{E} \big[ f(\overline{\mathbf{w}}^k) \big] - f^* \leq \epsilon $
		is satisfied if $k\geq \frac{\nu}{\epsilon} - \gamma$. That is, 
		\begin{align}
			K(\epsilon) = \frac{\nu}{\epsilon} - \gamma. \label{eq:n_iter_req}
		\end{align}
	\end{corollary}
	\begin{IEEEproof}
		The condition $k \geq \frac{\nu}{\epsilon} - \gamma$ can be rewritten as
		\begin{align} \label{eq:ineq_epsilon}
			\frac{\nu}{\gamma + k} \leq \epsilon.
		\end{align}
		Hence, from \eqref{ineq:thm_1} and \eqref{eq:ineq_epsilon}, $\mathbb{E} \big[ f(\overline{\mathbf{w}}^k) \big] - f^* \leq \epsilon$ holds for any $k 
		\geq 1$, which completes the proof.
	\end{IEEEproof}
	
	From Corollary \ref{col:n_iter_req},  we obtain the number of required 
	iterations $K(\epsilon)$ represented as \eqref{eq:n_iter_req}.
	
	\section{Analysis on Iteration Time} \label{sec:iter_time}
	In Section \ref{sec:required iterations}, we have characterized the number of  required iterations $K(\epsilon)$ as a function of $\epsilon$. To specify 
	the expected completion time in \eqref{eq:min_expected_time}, the expected time duration required for each iteration should 
	be analyzed. Moreover, as the randomness from data selection and communication is identical to every iteration, the expected 
	iteration time is the same. Thus, to enhance readability and avoid confusion, we omit the iteration index $k$ from here on.
	
	As stated in Section \ref{subsec:min_completion_time}, the required time duration for local updates is straightforwardly given as \eqref{eq:t_comp}. 
	Hence, after $\taucomp_n$, device $n$ is available for transmitting its updated local model. We focus on the analysis of communication time for 
	updated model transmission in this section. As briefly mentioned in Section \ref{subsubsec:comm_time}, this paper explores exemplary 
	communication models of both centralized and distributed, namely TDMA and RA, respectively. In the following two subsections, we investigate the 
	iteration time for each multiple access scheme. 
	
	\subsection{Iteration Time for TDMA}
	
	For TDMA, multiple devices are scheduled to transmit in different time slots. Suppose that, at a certain time slot, only one device has completed its 
	local updates. In such a case, it's evident that the device should transmit during that time. However, if multiple devices are ready for transmission, 
	without loss of generality, we assume that the server prioritizes scheduling the device with the lower index first. Note that 
	device indices are assigned in ascending order according to the batch size. Consequently, the time slot when device $n$ is scheduled to transmit is 
	given as
	\begin{align}
		\tau^{\text{TDMA}}_n = \max \left\lbrace \taucomp_n , \tau^{\text{TDMA}}_{n-1} \right\rbrace + 1, \label{eq:tcomm_TDMA}
	\end{align} where $\tau^{\text{TDMA}}_{n-1}$ is the time slot when device $n-1$ is scheduled to transmit and $\tau^{\text{TDMA}}_{0}=0$.
	
	Since the iteration time is defined as the time duration until the last device has delivered its local model to the server, the iteration time is the same as 
	the time slot scheduled for the device $N$ which has the largest batch size. Therefore,
	\begin{align}
		T^{\text{iter, TDMA}} &= \tau_N^{\text{TDMA}} , \nonumber\\
		& = \max \left\lbrace \taucomp_N , \tau^{\text{TDMA}}_{N-1} \right\rbrace + 1 \nonumber\\
		& = \max \left\lbrace \taucomp_N, \max\left\lbrace \taucomp_{N-1} , \tau^{\text{TDMA}}_{N-2} \right\rbrace + 1  \right\rbrace + 1 \nonumber\\
		&= \max \left\lbrace \taucomp_N, \taucomp_{N-1} +1 , \tau^{\text{TDMA}}_{N-2} + 1 \right\rbrace  + 1 ,
	\end{align}
	where the last equality holds since $\max \left\lbrace A, \max \left\lbrace B, C \right\rbrace \right\rbrace = \max \left\lbrace A, B, C \right\rbrace $ for 
	any $A$, $B$, and $C$. Then, by applying \eqref{eq:tcomm_TDMA} recursively, the iteration time using TDMA can be rewritten as 
	\begin{align} 
		T^{\text{iter, TDMA}} & =\max \left\lbrace \taucomp_N, \taucomp_{N-1} +1 , \cdots , \taucomp_1 + N-1 \right\rbrace + 1. \label{eq:t_iter_TDMA}
	\end{align}
	

	\subsection{Iteration Time for RA}	\label{subsec:iteration_time_RA}
	In the case of RA, we assume a simple RA protocol where each device decides whether to transmit or not using a certain probability known as the 
	transmission probability, denoted by $\ptr\in(0,1]$. As a result, multiple devices may transmit during the same time slot, leading to potential collisions. 
	In essence, successful transmission takes place only if one of the devices transmits while the others remain silent. In the event of a collision, 
	retransmission of the updated model becomes necessary until the server successfully receives it. Consequently, the probability of successful 
	transmission when $M\in[1:N]$ devices are available for transmission is expressed as
	\begin{align}
		\psuc_{M} = M (1 - \ptr)^{M-1} \ptr . \label{eq:psuc}
	\end{align}
	
	As seen in \eqref{eq:psuc}, successful transmission is probabilistically determined and, as a consequence, the number of time slots used for 
	delivering updated models is also stochastically given. Hence, we consider the expected iteration time denoted as $\mathbb{E}[T^{\text{iter, 
	RA}}]$. 	
	
	Note that the devices finishing local updates attempt transmission of their updated models to the server.  Hence, device $N$, which is allocated the 
	largest batch size, can send updated model after $\taucomp_N$. Moreover, due to collision, some devices which have completed updating earlier 
	than device $N$ may fail to deliver their updated models until time slot $\taucomp_N$. Hence, device $N$ may compete for channel access with 
	other devices that have not successfully transmitted updated models until time slot $\taucomp_N$. Let us denote $\navail\in[1:N]$ as the number of 
	devices that are required to transmit their updated models at time slot $\taucomp_N + 1$. Denote $\tcomm$ as the communication time duration in 
	which $\navail$ devices have successfully delivered their updated models, which is measured from time slot $\taucomp_N+1$. Then, the expected 
	iteration time for RA is represented by
	\begin{align} \label{eq:T_iter_RA}
		\mathbb{E} \left[ \titerra \right] = \taucomp_N + \mathbb{E} \left[ \tcomm \right] .
	\end{align} 
	
	As $\navail$ devices are attempting for transmission, $\tcomm$ can be expressed as a sum of $\navail$ inter-delivery time defined as the time 
	duration between consecutive successful deliveries. Let $\tcomm_m$ be the $m$th inter-delivery time, measured by the time duration between the 
	$(m-1)$th successful transmission and the $m$th successful transmission. We have 
	\begin{align} \label{eq:tcomm_m}
		\tcomm = \sum_{m=1}^{\navail} \tcomm_m.
	\end{align} 
	
	Considering that the devices attempt transmission until being successful, $\tcomm_m$ follows the geometric distribution. 
	Provided that $\navail$ devices are available at time slot $\taucomp_N +1$, $\tcomm_1$ is geometrically distributed with a probability 
	$\psuc_{\navail}$. After the first successful transmission, the second inter-delivery time is measured with the number of available devices 
	$\navail-1$, which means the probability of successful transmission is changed to $\psuc_{\navail-1}$. Consequently, the conditional probability 
	mass function (PMF) for the $m$th inter-delivery time given that $\navail$ devices are available at time slot $\taucomp_N+1$ is written as
	\begin{align}
		p_{\tcomm_m|\navail}(\tau) = (1 - \psuc_{\navail-m+1})^{\tau-1} \psuc_{\navail-m+1}, \label{eq:pmf_tcomm}
	\end{align}
	where $\tau=1,2,\cdots.$ Then, we have
	\begin{align} \label{eq:Ex_T_comm}
		\mathbb{E} [\tcomm ] &\overset{(a)}{=} \mathbb{E}_{\navail} \left[ \mathbb{E} [\tcomm | \navail] \right]  \nonumber\\
		& \overset{(b)}{=}  \mathbb{E}_{\navail} \left[  \mathbb{E} \left[ \sum_{m=1}^{\navail} \tcomm_m | \navail \right] \right ] \nonumber\\
		& \overset{(c)}{=}  \mathbb{E}_{\navail} \left[ \sum_{m=1}^{\navail} \frac{1}{\psuc_{\navail-m+1}} \right] \nonumber\\
		& = \mathbb{E}_{\navail} \left[ \sum_{m=1}^{\navail} \frac{1}{\psuc_m}\right],
	\end{align}
	where $(a)$ holds from the total probability, $(b)$ holds from \eqref{eq:tcomm_m}, $(c)$ holds from \eqref{eq:pmf_tcomm} and the fact that
	\begin{align}
		\mathbb{E} [ \tcomm_m | \navail] = \frac{1}{\psuc_{\navail-m+1}}.
	\end{align}

	Therefore, from \eqref{eq:T_iter_RA} and \eqref{eq:Ex_T_comm}
	\begin{align}
		\mathbb{E} [ \titerra ] = \taucomp_N + \mathbb{E}_{\navail} \left[ \sum_{m=1}^{\navail} \frac{1}{\psuc_m} \right] \label{eq:t_iter}.
	\end{align}
	
	Next, we will first investigate the distribution of $\navail$ for $N=2$ and $N=3$, and then consider arbitrary values of $N$.
	
	\subsubsection{Two-devices case}
	Let us consider the case where $N=2$. Then, $\navail$ becomes either one or two. First, we derive the probability that $\navail=1$. The event 
	$\navail=1$ means that device $1$ has succeeded in transmission before time slot $\taucomp_2+1$. Hence 
	\begin{align} \label{eq:p_N_avail_1}
		\Pr [ N^{\text{avail}} = 1 ] &= 1 - (1 -\psuc_1)^{\taucomp_2 - \taucomp_1}.
	\end{align}
	Obviously,
	\begin{align} \label{eq:p_N_avail_2}
		\Pr [ \navail = 2] & = 1 - \Pr[\navail=1] \nonumber\\
		& = (1 - \psuc_1)^{\taucomp_2 - \taucomp_1}.
	\end{align}
	
	Then, from \eqref{eq:t_iter}, \eqref{eq:p_N_avail_1}, and \eqref{eq:p_N_avail_2}, we have
	\begin{align}
		&\mathbb{E} [ \titerra ]\nonumber\\
		&= \taucomp_2 + \Pr[\navail = 1] \frac{1}{\psuc_1} + \Pr[\navail=2] \left( \frac{1}{\psuc_1} + \frac{1}{\psuc_2} \right) \nonumber\\
		& = \taucomp_2 + \frac{1}{\psuc_1} + \frac{1}{\psuc_2} (1 - \psuc_1)^{\taucomp_2 - \taucomp_1}  . \label{eq:t_iter_n_2}
	\end{align}
	
	\subsubsection{Three-devices case}
	For $N=3$, $\navail$ can be $1,2,$ or $3$. Obviously,
	\begin{align}
		\Pr [ N^{\text{avail}} = 1] = 1 - \Pr [ N^{\text{avail}} = 2] - \Pr [ N^{\text{avail}} = 3]. \label{eq:n1}
	\end{align}
	The event $\navail = 2 $ implies that one of the devices has succeeded in transmission before time slot $\taucomp_3 + 1$. Denote the time slot of 
	the first successful delivery by $\taucomm_1$. Then, depending on the value of $\taucomm_1$, we can derive joint probabilities as follows.
	
	Consider an event $\left\lbrace \navail = 2 , \taucomp_1 +1 \leq \taucomm_1 \leq \taucomp_2\right\rbrace$. When $\taucomp_1 +1 \leq \taucomm_1 
	\leq \taucomp_2$, the event $\navail = 2$ means that all communication attempts are failed during the time slots $[\taucomp_2 +1 , \taucomp_3]$. 
	Thus,
	\begin{align}
		&\Pr [ \navail = 2 , \taucomp_1+1 \leq \taucomm_1 \leq \taucomp_2 ] \nonumber\\
		&= (1 - (1 - \psuc_1)^{\taucomp_2 - \taucomp_1}) (1 - \psuc_1)^{\taucomp_3 - \taucomp_2}. \label{eq:n2_1}
	\end{align}		 
	On the other hand, if $\taucomp_2+1 \leq \taucomm_1 \leq \taucomp_3$, the first successful delivery should occur when device $1$ and device $2$ 
	are competing each other. In this case, after the first successful delivery, the remaining device is still attempting to transmit. To have $\navail =2$ for 
	this case, the remaining device fails in delivery during the time slots between $\taucomm_1 + 1$ and $\taucomp_3$. Therefore, $\Pr [\navail =2 , 
	\taucomp_2+1 \leq \taucomm_1 \leq \taucomp_3]$ is expressed as in \eqref{eq:n2_2}.
	\begin{figure*}[!htb]
		\begin{align}
			\Pr [\navail =2 , \taucomp_2+1 \leq \taucomm_1 \leq \taucomp_3]&= (1 - \psuc_1)^{\taucomp_2 - \taucomp_1} \sum_{\tau = 1}^{\taucomp_3 - 
			\taucomp_2} \psuc_2 ( 1- \psuc_2)^{\tau - 1} (1 - \psuc_1)^{\taucomp_3 - \taucomp_2 - \tau} \nonumber \\
			& = (1 - \psuc_1)^{\taucomp_2 - \taucomp_1} \psuc_2 \frac{ (1 - \psuc_1)^{\taucomp_3 - \taucomp_2 }}{1 - \psuc_2} \sum_{\tau = 
			1}^{\taucomp_3 - \taucomp_2} \left( \frac{ 1 - \psuc_2}{1 - \psuc_1} \right)^{\tau} \nonumber \\
			& = (1 - \psuc_1)^{\taucomp_2 - \taucomp_1} \frac{\psuc_2}{\psuc_2 -\psuc_1} \left( (1 - \psuc_1)^{\taucomp_3- \taucomp_2} - (1 - 
			\psuc_2)^{\taucomp_3 - \taucomp_2} \right) .
			\label{eq:n2_2}
		\end{align}
		\hrulefill
	\end{figure*}
	Because two events $\left\lbrace \taucomp_1+1 \leq \taucomm_1 \leq \taucomp_2 \right\rbrace$ and $ \left\lbrace \taucomp_2+1 \leq \taucomm_1 
	\leq \taucomp_3 \right\rbrace$ divides the sample space, 
	using \eqref{eq:n2_1} and \eqref{eq:n2_2}, $\Pr[ \navail = 2]$ is written as \eqref{eq:n2}.
	\begin{figure*}[!htb]
		\begin{align}
			\Pr [ \navail = 2] &= \Pr [ \navail = 2 , \taucomp_1+1 \leq \taucomm_1 \leq \taucomp_2 ]
			+ \Pr [\navail =2 , \taucomp_2+1 \leq \taucomm_1 \leq \taucomp_3] \nonumber\\
			\nonumber & = (1 - (1 - \psuc_1)^{\taucomp_2 - \taucomp_1}) (1 - \psuc_1)^{\taucomp_3 - \taucomp_2} \\
			&{~~~} +(1 - \psuc_1)^{\taucomp_2 - \taucomp_1} \frac{\psuc_2}{\psuc_2 -\psuc_1} \left( (1 - \psuc_1)^{\taucomp_3- \taucomp_2} - (1 - 
			\psuc_2)^{\taucomp_3 - \taucomp_2} \right)\nonumber\\
			& =(1 - \psuc_1)^{\taucomp_3 - \taucomp_2} +  \frac{\psuc_1}{\psuc_2 -\psuc_1}   (1 - \psuc_1)^{\taucomp_3- \taucomp_1}  - 
			\frac{\psuc_2}{\psuc_2 -\psuc_1} (1 - \psuc_1)^{\taucomp_2 - \taucomp_1}  (1 - \psuc_2)^{\taucomp_3 - \taucomp_2} .\label{eq:n2}
		\end{align} 
		\hrulefill
	\end{figure*}
	
	When $\navail = 3$, all communication attempts are failed before time slot $\taucomp_3+1$. Hence,
	\begin{align}
		\Pr [\navail =3] &= (1 - \psuc_1)^{\taucomp_2 - \taucomp_1} (1 - \psuc_2)^{\taucomp_3 - \taucomp_2}. \label{eq:n3}
	\end{align}
	By replacing $\Pr[\navail = 2]$ and $\Pr [ \navail = 3]$ in \eqref{eq:n1} with \eqref{eq:n2} and \eqref{eq:n3}, we have
	\begin{align} \label{eq:n1_2} 
		&\Pr [ \navail = 1] \nonumber\\
		&= 1 - \left( (1 - \psuc_1)^{\taucomp_3 - \taucomp_2} + \frac{\psuc_1}{\psuc_2 - \psuc_1} (1 -\psuc_1)^{\taucomp_3 - \taucomp_1} \right. 
		\nonumber\\
		\ & {~~~~} \left. - \frac{\psuc_1}{\psuc_2 - \psuc_1} (1 - \psuc_1)^{\taucomp_2 - \taucomp_1}(1 - \psuc_2)^{\taucomp_3 - \taucomp_2} \right) . 
	\end{align}
	Using the PMF of $\navail$ expressed in \eqref{eq:n2}, \eqref{eq:n3} and \eqref{eq:n1_2}, we obtain $\mathbb{E}\left[ \titerra \right]$ for $N=3$ as in 
	\eqref{eq:t_iter_n_3}.
	\begin{figure*}[!htb]
		\begin{align} 
			\mathbb{E}\left[ \titerra \right] 
			&= \taucomp_3 + \Pr[\navail = 1] \frac{1}{\psuc_1} + \Pr [ \navail = 2] \left( \frac{1}{\psuc_1} + \frac{1}{\psuc_2} \right) + \Pr [\navail = 3] \left( 
			\frac{1}{\psuc_1}+ \frac{1}{\psuc_2} + \frac{1}{\psuc_3} \right) \nonumber \\
			&=\taucomp_3 + \frac{1}{\psuc_1} + \frac{1}{\psuc_2} \left( (1 - \psuc_1)^{\taucomp_3 - \taucomp_2} + \frac{\psuc_1}{\psuc_2 - \psuc_1} (1 
			-\psuc_1)^{\taucomp_3 - \taucomp_1} \right.    \nonumber \\
			& {~~~}\left. - \frac{\psuc_1}{\psuc_2 - \psuc_1} (1 - \psuc_1)^{\taucomp_2 - \taucomp_1}(1 - \psuc_2)^{\taucomp_3 - \taucomp_2} \right)  + 
			\frac{1}{\psuc_3} (1 - \psuc_1)^{\taucomp_2 - \taucomp_1} (1 - \psuc_2)^{\taucomp_3 - \taucomp_2}. \label{eq:t_iter_n_3}
		\end{align} 
		\hrulefill
	\end{figure*}

	\subsubsection{N-devices case}
	For a general case with $N$ devices, the possible range of values for $\navail$ is given by $\left\lbrace 1, 2, \dots, N \right\rbrace$.
	In this case, we need to consider $N-1$ instances of the successful delivery time. Furthermore, the joint probability of $\navail$ and $\left\lbrace 
	\taucomm_n \right\rbrace_{n=1}^{N-1}$ varies depending on the range of each successful delivery. 
	Additionally, each sequence of $N-1$ successful deliveries has different probabilities due to the varying number of available devices. As an example, 
	when $N=5$, there exist $12$ distinct cases for $\navail = 1$. Consequently, addressing the complete PMF of $\navail$ and computing $\Pr [ \navail 
	= m]$ for all $m \in \left\lbrace 1, 2, \dots, N\right\rbrace$ involves considering numerous cases. As a result, obtaining a closed-form expression for 
	the PMF of $\navail$ becomes intractable. Instead, we can numerically derive the distribution of $\navail$ for a general $N$-device system using 
	Monte Carlo simulations.

	\section{Optimal Batch Allocation} \label{sec:opt_batch}
	
	In Section \ref{sec:iter_time}, we have analyzed the iteration time for both TDMA and RA protocols when the set of batch sizes $\{B_n\}_{n=1}^N$ 
	are given. In this section, we focus on an optimal batch allocation of $\{B_n\}_{n=1}^N$ under the constraint that $\sum_{n=1}^N B_n = B$.
	
	\subsection{Completion Time}
	From \eqref{eq:n_iter_req} and \eqref{eq:t_iter_TDMA}, we can rewrite the completion time $T$ in \eqref{eq:compl_t_tdma} for TDMA in 
	\eqref{eq:compl_t_tdma} as
	\begin{align}
		T = \left( \frac{\nu}{\epsilon} - \gamma \right) \left( \max \left\lbrace \taucomp_N, \taucomp_{N-1} +1 , \cdots , \taucomp_1 + N-1 \right\rbrace + 1 
		\right) ,\label{eq:t_completion_tdma} 
	\end{align}
	For the case of RA, the expected completion time is given by 
	\begin{align}
		\mathbb{E}[T] &= \left( \frac{\nu}{\epsilon} - \gamma \right) \left( \taucomp_N + \mathbb{E}_{\navail} \left[ \sum_{m=1}^{\navail} \frac{1}{\psuc_m} 
		\right] \right) \label{eq:t_completion_ra}
	\end{align}
	from \eqref{eq:n_iter_req} and \eqref{eq:t_iter}.
	
	Because the sum of batch size $B$ is a constant, the number of iterations required to achieve a target optimality gap becomes also a constant. Thus, 
	to minimize the completion time or the expected completion time, we need to minimize the iteration time for each iteration procedure, which depends 
	on $\{B_n\}_{n=1}^N$ for both TDMA and RA. To minimize waiting  or retransmission time for uploading locally updated models, we propose a batch 
	allocation algorithm called the step-wise batch allocation.  The pseudo code of the proposed algorithm is stated in Algorithm \ref{alg:sw_batch_allo}.

	\begin{algorithm}[pt]\caption{Step-wise batch allocation} \label{alg:sw_batch_allo}
		\begin{algorithmic}[1]
			\State {\bf Input} : $\Delta\in(0,B]$.
			\State {\bf Initialization}: Set $B^{\text{SW}}_n=0$ for all $n\in[1:N]$, $N'=0$, $B'=0$, and $\mbox{Flag}=\text{False}$.
			\While {$\mbox{Flag}==\text{False}$}	
			\For {$n\in[0:N']$} 
			\State $B^{\text{SW}}_{N-n}\leftarrow B^{\text{SW}}_{N-n} + \Delta $.
			\State  $B'\leftarrow B'+\Delta $.
			\If {$B'\geq B$}
			\State  $\mbox{Flag}\leftarrow \text{True} $.
			\If {$B' - B > 0 $}
			\State $B^{\text{SW}}_{N-n} \leftarrow B^{\text{SW}}_{N-n} - \left( B' - B \right)$
			\EndIf 
			\EndIf
			\State  $N' \leftarrow \min(N' + 1,N-1)$
			\EndFor
			\EndWhile
			\State {\bf Output}: $\{B^{\text{SW}}_n\}_{n=1}^N$.
		\end{algorithmic}
	\end{algorithm} 
	
	In Section \ref{subsec:batch_TDMA}, we prove that  Algorithm \ref{alg:sw_batch_allo} with $\Delta=\rho$ provides an optimal batch allocation 
	minimizing $T^{\text{iter, TDMA}} $ in \eqref{eq:t_iter_TDMA} for TDMA. In Section \ref{subsec:batch_RA}, we analyze an optimal batch allocation 
	for RA when $N=2$ and $N=3$. For $N>3$, we numerically optimize $\Delta\in(0,B]$ in Algorithm \ref{alg:sw_batch_allo} to minimize the iteration 
	time for RA in Section~\ref{sec:exp}.

	\subsection{Batch Allocation for TDMA} \label{subsec:batch_TDMA}
	
	In the following, we will prove that the output of the step-wise batch allocation achieves the minimum iteration time for TDMA.
	
	To show the optimality of the proposed step-wise batch allocation, we first derive a lower bound on $T^{\text{iter, TDMA}}$ when the total batch 
	size $B$ is given.
	
	\begin{proposition}\label{prop_titer_LB}
		Suppose that there exists a positive integer $m\geq 1$ satisfying that $\rho \left( (m-1)N + \frac{N(N+1)}{2} \right) < B \leq \rho \left( m N + 
		\frac{N(N+1)}{2} \right)$. Then the iteration time for TDMA is lower-bounded by
		\begin{align}
			T^{\text{iter, TDMA}} \geq m + N+1.
		\end{align}
	\end{proposition}
	\begin{IEEEproof}
		We prove this proposition by contradiction. For such purpose, assume that $T^{\text{iter, TDMA}} < m+ N+1$ 
		is achievable. From \eqref{eq:tcomm_TDMA}, every device should be scheduled no later 
		than $m + n$. That 
		means  
		\begin{align}
			\tau^{\text{TDMA}}_n \leq m + n \label{eq:thm2_b1}
		\end{align}
		for $n\in[1:N]$.
		Moreover, in federate learning, only the devices that have finished their local updates can participate in transmission. Thus,
		\begin{align}
			\taucomp_n \leq \tau^{\text{TDMA}}_n - 1 . \label{eq:thm2_b2}
		\end{align}
		Combining \eqref{eq:thm2_b1} and \eqref{eq:thm2_b2}, we have
		\begin{align}
			\taucomp_n \leq m + n - 1. \label{eq:thm2_b3}
		\end{align}
		Then, from \eqref{eq:t_comp} and \eqref{eq:thm2_b3} , we have
		\begin{align}
			B_n \leq \rho \left( m + n - 1 \right). \label{eq:thm2_b4}
		\end{align}
		Since $B = \sum_{n=1}^N B_n$, we can find an upper bound on $B$ using \eqref{eq:thm2_b4}.
		\begin{align}
			B &\leq \sum_{n=1}^N \rho \left( m + n - 1 \right) , \\
			& \leq \rho \left( (m-1) N + \frac{N(N+1)}{2} \right) . \label{eq:thm3_b5}
		\end{align}
		However, \eqref{eq:thm3_b5} contradicts to the assumption that $B > \rho \left( (m-1)N + \frac{N(N+1)}{2} \right)$. As a consequence, 
		$T^{\text{iter, TDMA}} \geq m + N+1$, which completes the proof.
	\end{IEEEproof}
	
	By using Proposition \ref{prop_titer_LB}, we can prove the optimality of the proposed step-wise batch allocation algorithm, stated in Theorem 
	\ref{thm:min_titer_TDMA}.	
	
	\begin{theorem}\label{thm:min_titer_TDMA}
		Suppose that there exists a positive integer $m\geq 1$ satisfying that $\rho \left( (m-1)N + \frac{N(N+1)}{2} \right) < B \leq \rho \left( m N + 
		\frac{N(N+1)}{2} \right)$. Then the step-wise batch allocation $\{B^{\text{SW}}_n\}_{n=1}^N$ from Algorithm \ref{alg:sw_batch_allo} achieves 
		$T^{\text{iter, TDMA}} = m+N+1$ by setting $\Delta=\rho$.
	\end{theorem}
	\begin{IEEEproof}
		First, let us consider the case where $B = \rho \left( m N + \frac{N(N+1)}{2} \right)$. For this case, $\left\lbrace B_n^{\text{SW}} 
		\right\rbrace_{n=1}^N$ is given by
		\begin{align}
			B_n^{\text{SW}} = \rho ( m + n ) \label{eq:thm3_b_n}
		\end{align}
		for $n\in[1:N]$. 
		From \eqref{eq:thm3_b_n},
		\begin{align}
			B_n^{\text{SW}} - B_{n-1}^{\text{SW}} = \rho
		\end{align}
		and, as a result, we have 
		\begin{align}
			\taucomp_n = \taucomp_{n-1} + 1.
		\end{align}
		Therefore, each device $n$ can be scheduled one time slot earlier than the next device. Consequently, at the $(m + N)$th time slot, all devices 
		except for device $N$ have completed updating models and transmitted updated models to the server. Moreover, device $N$ finishes updating at 
		the $(m + N)$th time slot. As a consequence, it is possible for device $N$ to send its updated model at the $(m + N+1)$th time slot. That means 
		that the step-wise batch allocation algorithm can achieve $T^{\text{iter, TDMA}} = m + N+1$ for $B = \rho \left( m N + \frac{N(N+1)}{2} \right)$.
		
		For $ \rho \left( (m-1) N + \frac{N(N+1)}{2} \right) < B < \rho \left( m N + \frac{N(N+1)}{2} \right)$, the resulting batch allocation from Algorithm 
		\ref{alg:sw_batch_allo} becomes identical to that for $B = \rho \left( m N + \frac{N(N+1)}{2} \right)$ by removing some batch assigned in the later 
		loop. Since reducing the allocated batch size cannot increase the computation or communication time, $T^{\text{iter, TDMA}} = m + N+1 $ is still 
		achievable for this case. This completes the proof.
	\end{IEEEproof}
	
	
	\subsection{Batch Allocation for RA} \label{subsec:batch_RA}
	When RA is employed for transmitting updated models, the time duration for computation and communication affects the expected iteration time, as 
	demonstrated in \eqref{eq:t_iter}. If the server assigns equal-size batches to all devices, the time required for local updates is reduced by leveraging 
	parallel computing. However, after updating the models, all devices will simultaneously attempt transmission, leading to longer communication times 
	due to heavy contention.
	On the other hand, if there are variations in batch sizes allocated to devices, the time slot when the last device with the largest batch size completes 
	its updates can be delayed. However, devices with smaller batch sizes will have opportunities to transmit updated models while other devices are still 
	in the process of updating their models.
	
	Hence, it is crucial to allocate an appropriate batch size to each of $N$ devices. By combining  \eqref{eq:t_comp} and \eqref{eq:t_iter}, the expected 
	iteration time for RA is approximately given as $\mathbb{E} [ \titerra ] \simeq  \frac{B_N}{\rho} + \mathbb{E}_{\navail} \left[ \sum_{m=1}^{\navail} 
	\frac{1}{\psuc_m} \right]$, where we ignore the ceiling operation in \eqref{eq:t_comp}.
	Therefore, we aim to address the following optimization problem to minimize the expected iteration time for RA:
	
	\begin{align}
		\textbf{P1} : & \min_{\{B_n\}_{n=1}^N} \left(  \frac{B_N}{\rho} + \mathbb{E}_{\navail} \left[ \sum_{m=1}^{\navail} \frac{1}{\psuc_m} \right] \right)
	\end{align} 
	subject to
	\begin{align}
		\sum_{n=1}^N B_n = B.
	\end{align}
	Denote the solution of the optimization problem $\textbf{P1}$ by $\{B^*_n\}_{n=1}^N$.
	Note that, in order to solve $\textbf{P1}$, we need to know the distribution of $\navail$. In Section \ref{subsec:iteration_time_RA}, we established a 
	closed-form expression for the distribution of $\navail$ when $N=2$ and $N=3$. Based on these results, we derive the optimal batch allocation 
	$\{B^*_n\}_{n=1}^N$ when $N=2$ in the following theorem.
	
	\begin{theorem} \label{thm:optimal_N_2}
		For $N=2$, the optimal batch allocations in $\textbf{P1}$, denoted by $\{B_1^*,B_2^*\}$, are given as \eqref{eq:opt_b_1}.
		\begin{figure*}[!htb]
			\begin{align}
				\left\{B_1^*,B_2^*\right\} &= \begin{cases}
					\left\{\frac{B}{2},\frac{B}{2}\right\} &  \text{ if } \frac{q}{\ln (1 - \psuc_1)} \ln \left( - \frac{\psuc_2}{ 2 \ln (1 - \psuc_1)} \right) < 0, \\
					\{0,B\} &  \text{ if } \frac{q}{\ln (1 - \psuc_1)} \ln \left( - \frac{\psuc_2}{ 2\ln (1 - \psuc_1)} \right) > B,\\
					\left\{\frac{1}{2} \left( B - \frac{q}{\ln (1 - \psuc_1)} \ln \left( - \frac{\psuc_2}{2\ln (1 - \psuc_1)} \right) \right),\frac{1}{2} \left( B + \frac{q}{\ln (1 - 
					\psuc_1)} \ln \left( - \frac{\psuc_2}{ 2\ln (1 - \psuc_1)} \right) \right)\right\}  &  \text{ Otherwise.} 
				\end{cases}\label{eq:opt_b_1} 
			\end{align}
			\hrulefill
		\end{figure*}
	\end{theorem}
	\begin{IEEEproof}
		By defining the difference between $B_1$ and $B_2$ as $\Delta = B_2 - B_1$, we have $B_1 = \frac{B}{2} - \frac{\Delta}{2}$ and $B_2 = 
		\frac{B}{2} + \frac{\Delta}{2}$. Then from \eqref{eq:t_iter_n_2}, the original problem $\textbf{P1}$ can be rewritten as
		\begin{align}
			\textbf{P2} : & \min_{\Delta\in[0,B]} \left( \frac{\Delta}{2\rho} + \frac{1}{\psuc_2} (1- \psuc_1)^{\frac{\Delta}{\rho}} \right)\label{eq:obj_f_n_2}.
		\end{align}
		
		For notational convenience, denote the objective function in \eqref{eq:obj_f_n_2} by $g(\Delta)$. Then, the derivative of $g(\Delta)$ is given by 
		\begin{align}
			\frac{d g(\Delta)}{d \Delta} & = \frac{1}{2\rho} + \frac{\ln (1 - \psuc_1)}{\psuc_2 \rho} (1 - \psuc_1)^{\frac{\Delta}{\rho}} .
		\end{align}
		
		Suppose that $\frac{dg(\Delta)}{d\Delta} > 0 $ for $\Delta\in[0,B]$. Then, the completion time increases as $\Delta $ increases within the feasible 
		range of $[0,B]$. As a result, $\Delta^*=0$ provides the optimal batch allocation, given by $\{B_1^*, B_2^*\}=\left\{\frac{B}{2},\frac{B}{2}\right\}$. 
		Note that the condition $\frac{dg(\Delta)}{d\Delta} > 0 $ is rewritten as
		\begin{align}
			\frac{\rho}{\ln (1 - \psuc_1)} \ln \left( - \frac{\psuc_2}{2 \ln (1 - \psuc_1)} \right)< \Delta.
		\end{align}
		Hence, if $\frac{\rho}{\ln (1 - \psuc_1)} \ln \left( - \frac{\psuc_2}{\ln (1 - \psuc_1)} \right) < 0$, then $\frac{dg(\Delta)}{d\Delta} > 0$ holds for $0 \leq 
		\Delta \leq B$.
		
		On the other hand, if $\frac{dg(\Delta)}{d\Delta} < 0 $ for $\Delta\in[0,B]$, the completion time decreases as $\Delta$ increases. Hence,  
		$\Delta^*=B$ provides the optimal batch allocation, given by $\{B_1^*, B_2^*\}=\left\{0,B\right\}$ for this case. In a similar manner, if $\frac{\rho}{\ln 
		(1 - \psuc_1)} \ln \left( - \frac{\psuc_2}{ 2 \ln (1 - \psuc_1)} \right) > B $, $\frac{dg(\Delta)}{d\Delta} < 0 $ holds for $0 \leq \Delta \leq B$.
		
		Lastly, there is a case where neither $\frac{dg(\Delta)}{d\Delta} > 0 $ nor $\frac{dg(\Delta)}{d\Delta} < 0 $ holds for $\Delta\in[0,B]$. In that case, 
		the optimal condition corresponds to the point where the derivative is equal to zero. Hence, from the condition $ \frac{dg(\Delta)}{d\Delta} 
		\big|_{\Delta^* } = 0$, we have
		\begin{align}
			\Delta^* = \frac{\rho}{\ln (1 - \psuc_1)} \ln \left( - \frac{\psuc_2}{ 2 \ln (1 - \psuc_1)} \right).
		\end{align}
		Hence, $B_1^*=\frac{1}{2} \left( B - \frac{q}{\ln (1 - \psuc_1)} \ln \left( - \frac{\psuc_2}{ 2 \ln (1 - \psuc_1)} \right) \right)$ and $B_2^*=\frac{1}{2} \left( 
		B + \frac{q}{\ln (1 - \psuc_1)} \ln \left( - \frac{\psuc_2}{2 \ln (1 - \psuc_1)} \right) \right)$. This completes the proof.
	\end{IEEEproof}
	
	For $N=3$, $\textbf{P1}$ can be rewritten as \eqref{eq:obj_f_n_3} subject to
	\setcounter{equation}{74}
	\begin{align}
		2 \Delta_1 + \Delta_2 \in[0,B], \label{ineq:const} 
	\end{align}
	where $\Delta_1=B_2-B_1$ and $\Delta_2=B_3-B_2$.
	Since the optimization problem $\textbf{P3}$ is a convex programming problem, we can find an optimal solution using the Karush--Kuhn--Tucker 
	(KKT) conditions \cite{Boyd_book2004}. Let us denote the objective function in \eqref{eq:obj_f_n_3} as $h(\Delta_1 , \Delta_2)$ and the Lagrangian 
	function as 
	\begin{align}
		&\mathcal{L}(\Delta_1, \Delta_2, \alpha_1, \alpha_2, \alpha_3 ) \nonumber\\
		&= h(\Delta_1, \Delta_2) - \alpha_1 \Delta_1 - \alpha_2 \Delta_2 +  \alpha_3( 2 \Delta_1 
		+ \Delta_2  - B).		
	\end{align}
	Then, the corresponding KKT conditions are represented as
	\begin{align}
		\left. \frac{\partial \mathcal{L}}{\partial \Delta_i}\right|_{\Delta_i = \Delta_i^*} &= 0  \hspace{15pt}\mbox{for }   i =1,2, \label{eq:thm_3_6} \\
		\Delta_i^* &\geq 0  \hspace{15pt}\mbox{for }   i =1,2, \\
		\alpha_i &\geq 0  \hspace{15pt}\mbox{for }   i =1,2,3,  \\
		\alpha_i \Delta_i^* &= 0  \hspace{15pt}\mbox{for }   i =1,2,\\
		\alpha_3 ( 2 \Delta_1^* + \Delta_2^* - B ) &= 0.
	\end{align}
	
	\setcounter{equation}{73}
	\begin{figure*}[!htb]
		\begin{align}
			\textbf{P3} :  & \min_{\Delta_1, \Delta_2 } \left\lbrace \frac{\Delta_1 + \Delta_2}{\rho} + \frac{1}{\psuc_2} \left( (1 - \psuc_1)^{\frac{\Delta_2}{\rho}} 
			+ \frac{\psuc_1}{\psuc_2- \psuc_1 } (1 -\psuc_1)^{\frac{\Delta_1 + \Delta_2}{\rho} }  \right.  \right. \nonumber\\
			&  {~~~~~~~}\left. \left. - \frac{\psuc_1}{\psuc_2 - \psuc_1} (1 - \psuc_1)^{\frac{\Delta_1 }{\rho} }(1 - \psuc_2)^{\frac{\Delta_2}{\rho}} \right)  + 
			\frac{1}{\psuc_3} (1 - \psuc_1)^{\frac{\Delta_1 }{\rho} } (1 - \psuc_2)^{\frac{\Delta_2}{\rho}}  \right\rbrace \label{eq:obj_f_n_3} 
		\end{align} 
		\hrulefill
	\end{figure*}
	\setcounter{equation}{75}

	For $N>3$, obtaining a closed-form expression for the PMF of $\navail$ is intractable since it requires accounting for all combinations of $N-1$ 
	realizations of successful delivery times. However, the optimal solutions for $N=2$ and $N=3$ demonstrate that different sizes of batches between 
	the devices  is necessary to reduce the expected iteration time. Moreover, when TDMA is used, the step-wise batch allocation in Algorithm 
	\ref{alg:sw_batch_allo} with $\Delta=\rho$ is shown to be optimal. Hence, for $N>3$ we numerically optimize $\Delta\in(0,B]$ in Algorithm 
	\ref{alg:sw_batch_allo} to minimize the iteration time for RA in Section~\ref{sec:exp}. Unlike TDMA in which a single time slot is required for 
	transmitting the updated model of each device due to centralized scheduling, devices communicate stochastically for RA, which might necessitate a 
	larger number of time slots, i.e., $\Delta > \rho$.

	\section{Experiments}\label{sec:exp}
	In this section, we present experimental results demonstrating the completion time using real datasets.
	
	We evaluate the performance of the step-wise batch allocation in Algorithm \ref{alg:sw_batch_allo} based on the MNIST classification. For the sake 
	of simplicity, we concentrate on binary classification, using only the ``$0$'' and ``$8$'' classes from the MNIST dataset, as detailed in 
	\cite{Shi_ISIT2021}. Additionally, we employ the cross-entropy loss function. For this case, we have $L = 1$ and $M = 1$. The remaining parameters 
	are set as $\gamma = 1$, $c = 1.5$, $\lambda = 0.1$, and $B=10000$. 
	
\begin{figure}[pt]
		\centering
		\includegraphics[scale=0.45]{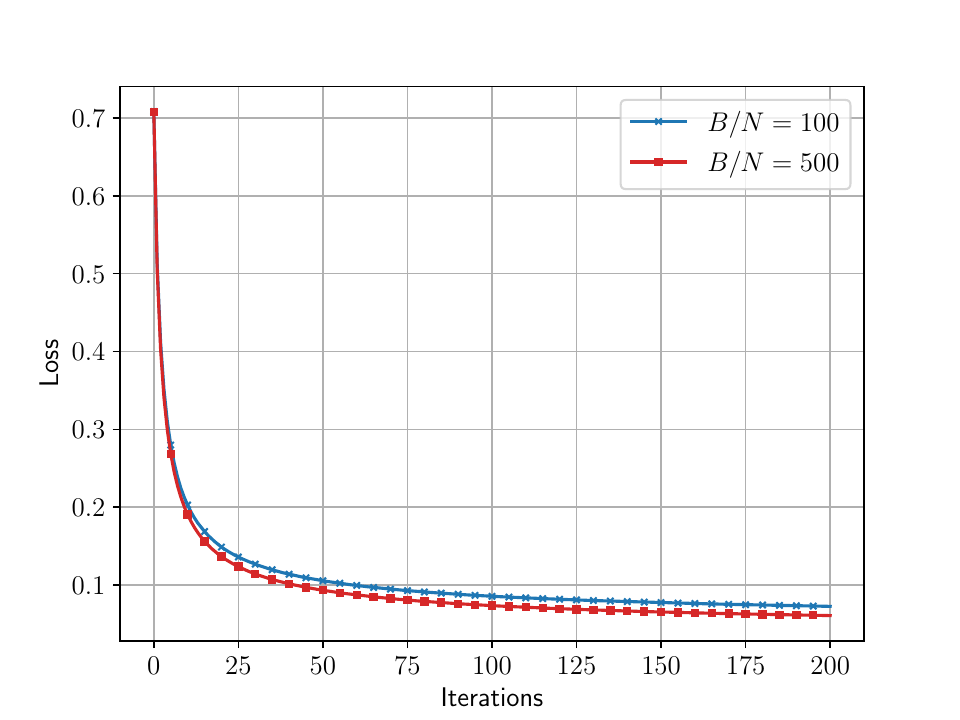}
		\caption{Training loss as a function of the number of iterations.}\label{fig_iter_loss}
	\end{figure}

	To validate the efficacy of $K(\epsilon)$ derived in Theorem \ref{thm_conv} and Corollary \ref{col:n_iter_req}, we assess the training loss in Fig. 
	\ref{fig_iter_loss}. The model undergoes training with an average batch size per device denoted as $B/N = 100, 500$ over a span of 200 iterations. 
	From Corollary \ref{col:n_iter_req}, the values of $K(0.1)$ and $K(0.05)$ are specified as $19$ and $39$ respectively for both cases. Notably, at the 
	$200$th iteration, the corresponding loss values are recorded as $0.727$ and $0.609$ for $\frac{B}{N} = 100, 500$, respectively. Treating the loss 
	value at the 200th iteration as an approximation of the minimum loss, the optimal gap $\epsilon = 0.1$ is attained at the $15$th iteration for both 
	cases, while $\epsilon = 0.05$ is achieved at the $33$rd iteration for $\frac{B}{N} = 100$ and at the $32$nd iteration for $\frac{B}{N}= 500$, 
	respectively.

	\begin{figure}[pt]
		\centering
		\includegraphics[scale=0.45]{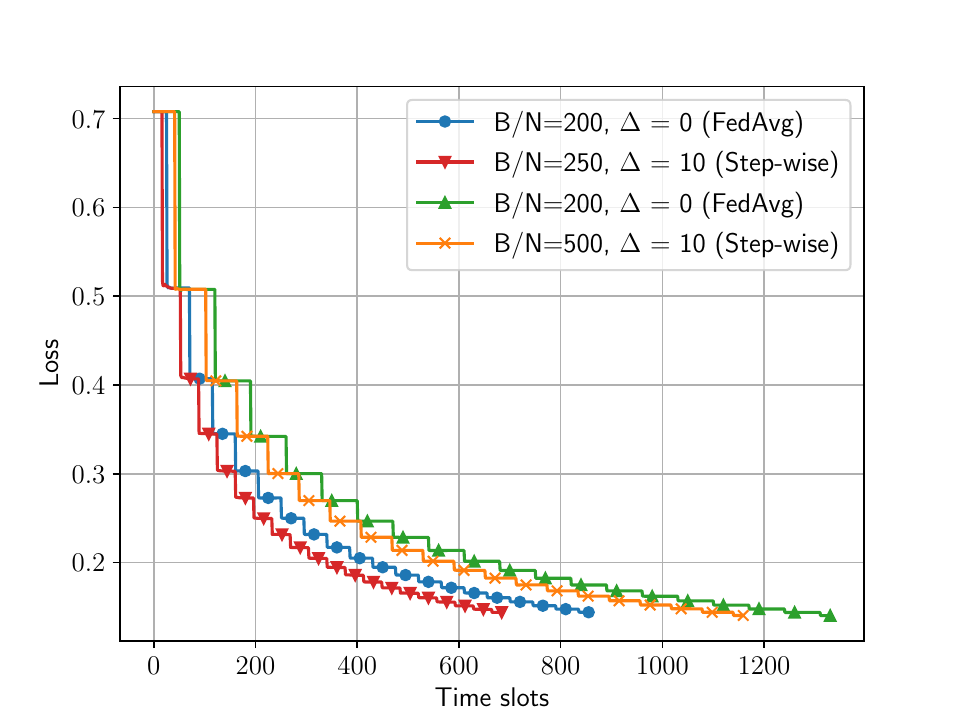}
		\caption{Comparison of global loss for different batch gap using TDMA for $N=20$, $\rho = 10$.}\label{fig_tdma_loss}
	\end{figure}
	
	\begin{figure}[pt]
		\centering
		\includegraphics[scale=0.45]{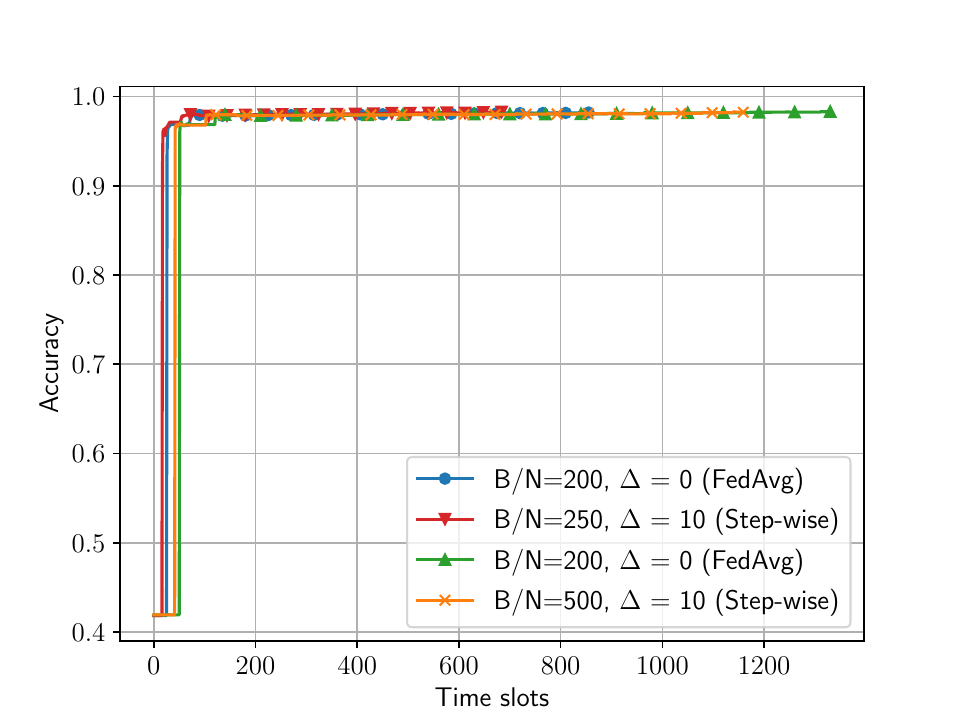}
		\caption{Comparison of accuracy for different batch gap using TDMA for $N=20$, $\rho = 10$.} \label{fig_tdma_acc}
	\end{figure}
	
	In Fig. \ref{fig_tdma_loss} and \ref{fig_tdma_acc}, we compare the performance of federated learning with different size of batch gap under the 
	TDMA protocol. We applied the step-wise batch allocation in Algorithm~\ref{alg:sw_batch_allo}. It is worth highlighting that federated learning with 
	$\Delta = 0$ is essentially equivalent to FedAvg \cite{McMahan_MLR2017}. As seen in the figures, the step-wise batch allocation with $\Delta = 10$ 
	exhibits faster convergence compared to FedAvg, which corresponds to the case where $\Delta = 0$. Indeed, Theorem \ref{thm:min_titer_TDMA} 
	proved that the proposed step-wise batch allocation with $\Delta=\rho$ minimizes the completion time.
	
\begin{figure}[pt]
		\centering
		\includegraphics[scale=0.45]{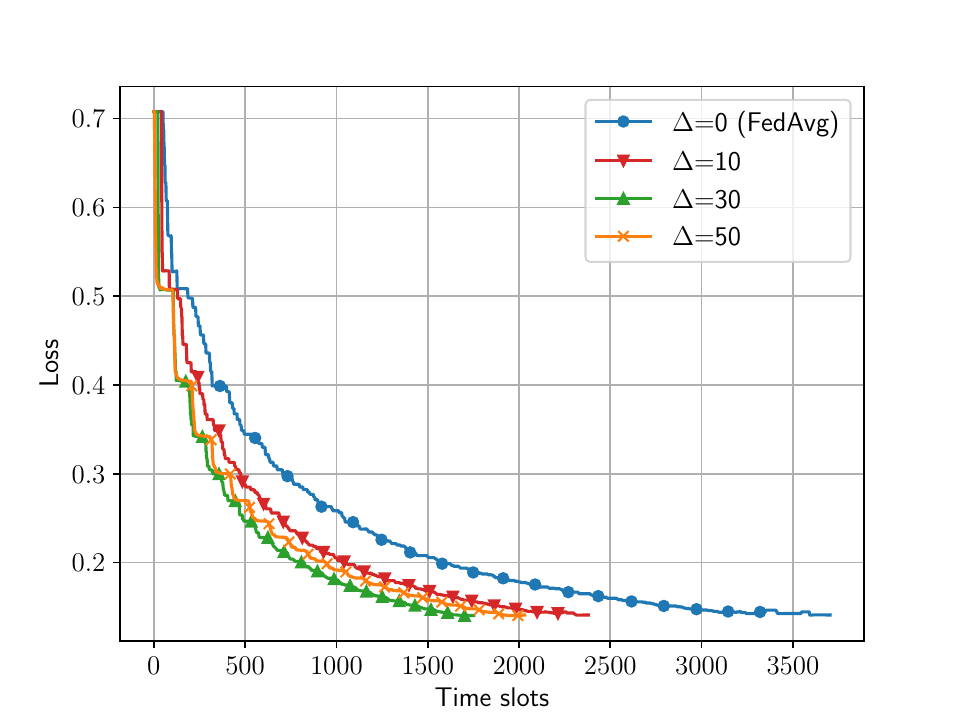}
		\caption{Comparison of global loss for different batch gap using RA for $N=20$, $\rho = 10$, $\ptr = 0.2$.} \label{fig_4}
	\end{figure}

	In Fig. \ref{fig_4}, we evaluate the value of loss function as a function of time slots using RA for $N=20$, $\rho = 10$, and $\ptr = 0.2$. 
	We again applied the step-wise batch allocation in Algorithm~\ref{alg:sw_batch_allo}.
	Although the communication protocol has been changed from TDMA to RA, models are updated and aggregated under the same rule using SGD. 
	Consequently, performance of learning does not change notably even after change of communication protocol. 	
	As seen in the figure, the step-wise batch allocation with $\Delta = 30$ shows fastest reduction of loss function compared with the case where 
	$\Delta = 0$.

\begin{figure}[pt]
		\centering
		\includegraphics[scale=0.45]{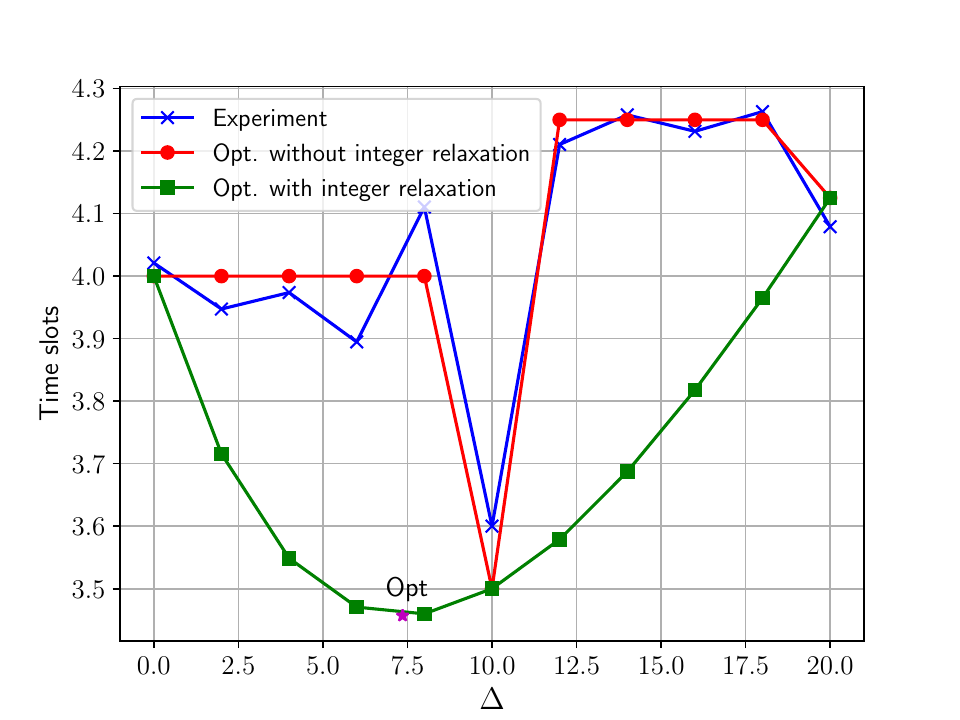}
		\caption{Iteration time for $N=2$, $B = 500$, $\rho=5$.}\label{fig_N_2}
	\end{figure}

	For $N=2$, the optimal batch allocation was determined in Theorem \ref{thm:optimal_N_2} by disregarding the ceiling operation. In Fig. 
	\ref{fig_N_2}, the gap between iteration time and average computing time, defined as $\frac{B}{\rho N}$, is compared with experimental results. In 
	the figure, `Opt. with integer relaxation' corresponds to the solution of \eqref{eq:obj_f_n_2}. Then, `Opt. without integer relaxation' is obtained by 
	applying the ceiling operation to `Opt. with integer relaxation'. Due to integer constraints, a gap from the theoretically optimal point obtained under 
	integer relaxation is inevitable. However, this gap can be considered negligible.
	
	\begin{figure}[pt]
		\centering
		\includegraphics[scale=0.45]{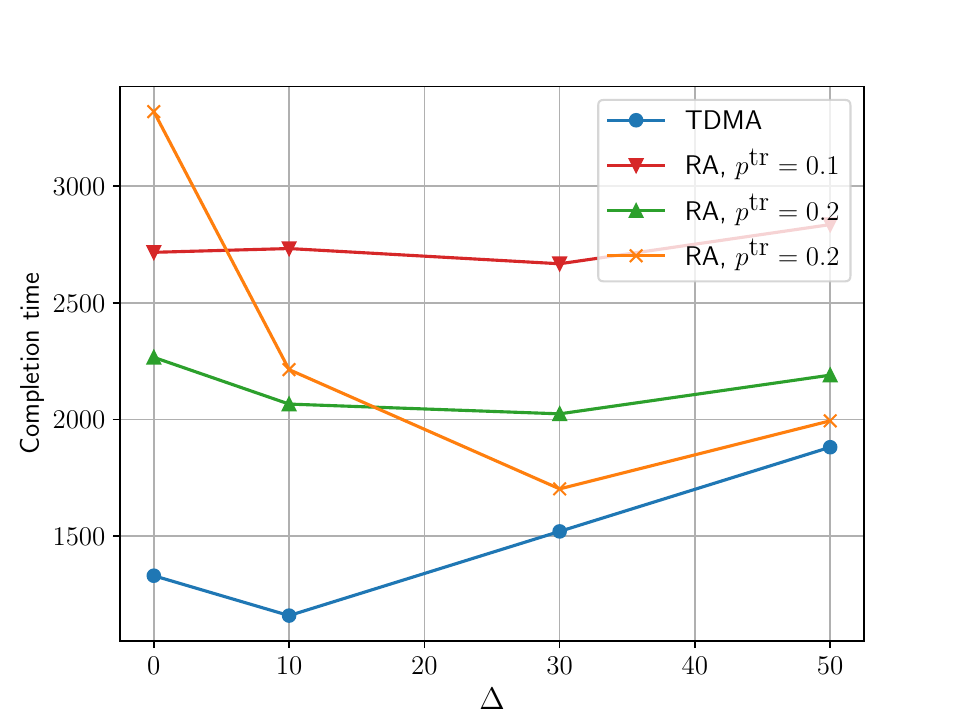}
		\caption{Comparison of completion time for different batch gap for $N=20$, $B=10000$.} \label{fig_9}
	\end{figure}
	
	Fig. \ref{fig_9} represents the completion time as a function of $\Delta$ for different values of the transmission probability $\ptr$ for RA. The 
	completion time of TDMA is also included. When TDMA is utilized, the server schedules transmissions in a way that avoids collisions, resulting in the 
	lowest completion time for federated learning. Moreover, as $\ptr$ for RA increases, the performance of step-wise batch allocation with an 
	appropriate batch gap achieves a lower completion time.
	It is known that the $\ptr$ that minimizes communication time for RA is given as the inverse of the number of nodes~\cite{Jeon_TMCOM2023}. 
	Thus, for $N=20$, the optimal transmission probability is determined as $\ptr = 0.05$. However, when comparing the completion times for $\ptr = 
	0.05$ and $\ptr = 0.2$ in Fig. \ref{fig_9}, the completion time for $\ptr = 0.2$ is smaller than that of $\ptr = 0.05$ by properly setting $\Delta$.

\begin{figure}[pt]
		\centering
		\includegraphics[scale=0.45]{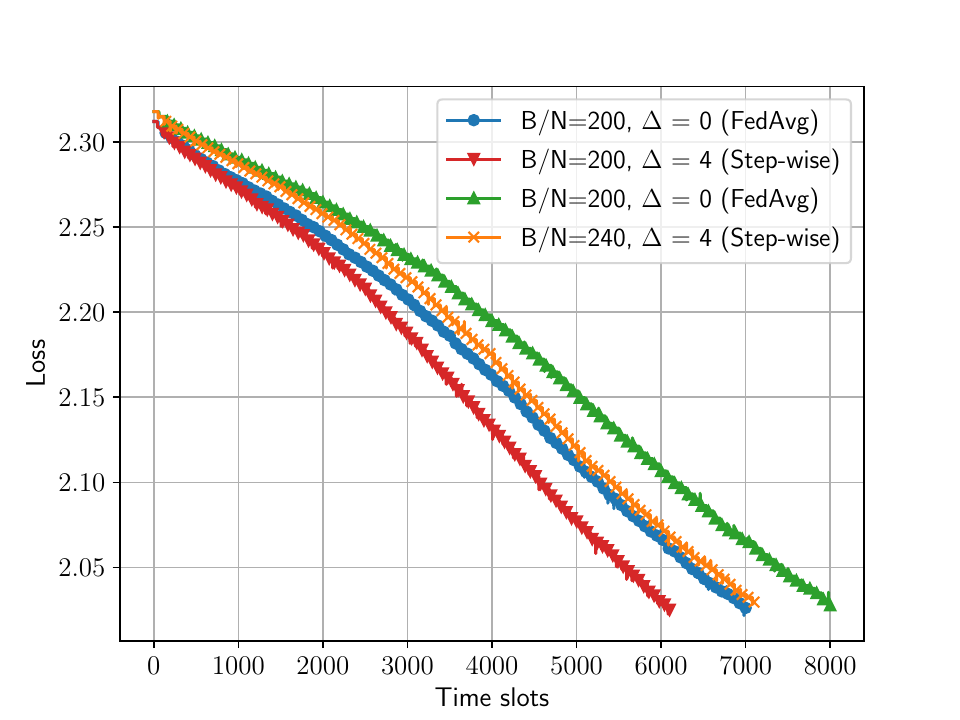}
		\caption{Comparison of loss function for $N=20$, $\rho = 4$ with CIFAR10 dataset using TDMA.} \label{fig_10}
	\end{figure}

\begin{figure}[pt]
		\centering
		\includegraphics[scale=0.45]{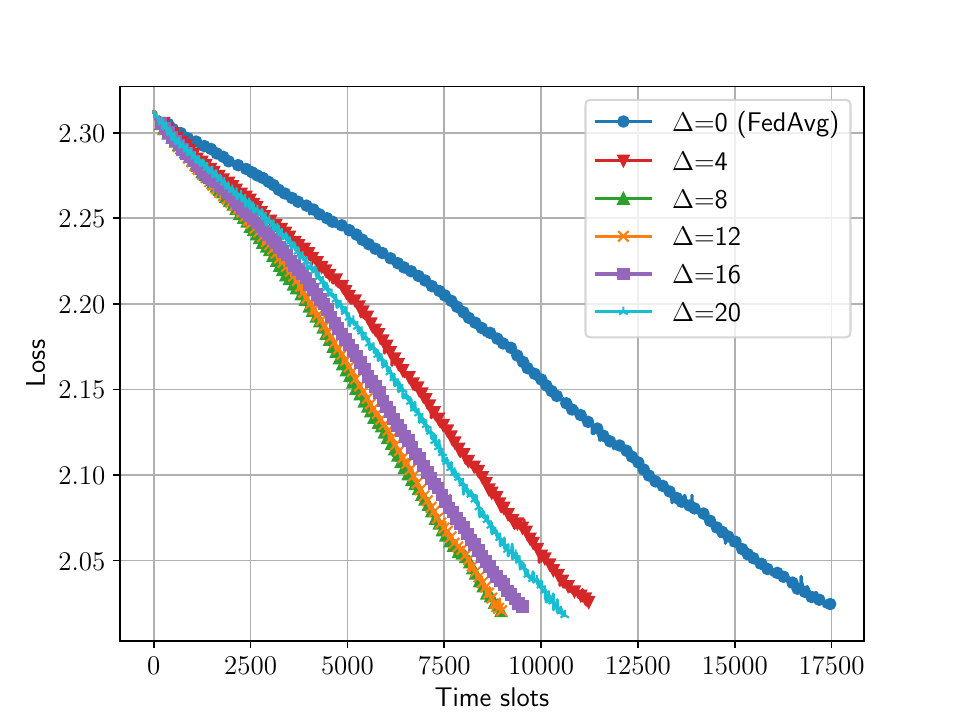}
		\caption{Comparison of loss function for $N=20$, $B=4000$, $\rho = 4$ with CIFAR10 dataset using RA when $\ptr = 0.2$.} \label{fig_11}
	\end{figure}

\begin{figure}[pt]
		\centering
		\includegraphics[scale=0.45]{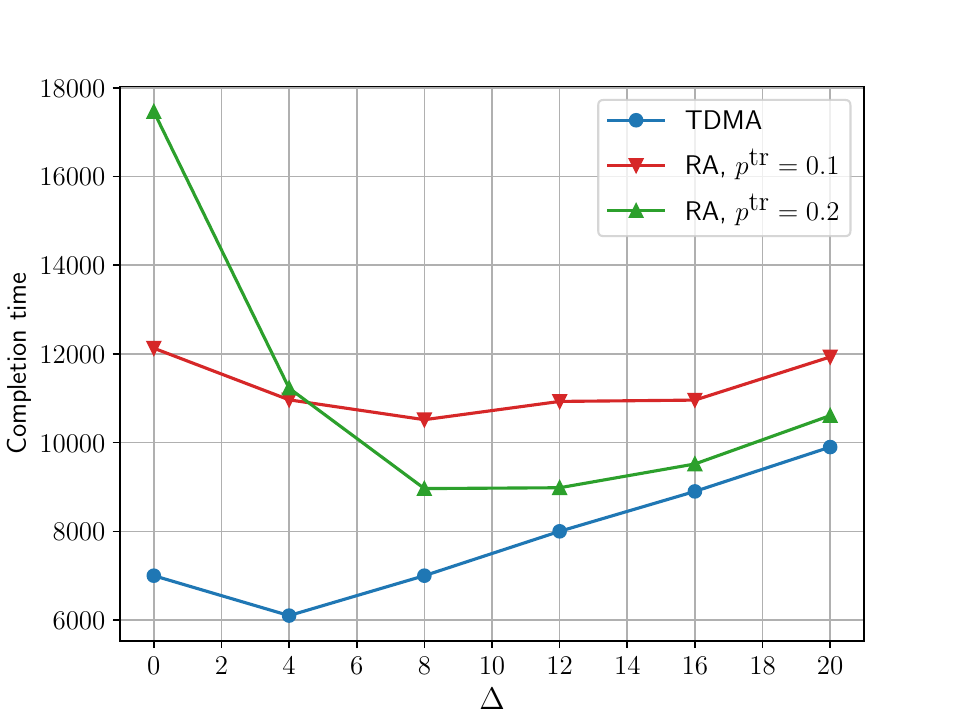}
		\caption{Comparison of completion time for different batch gap when $N=20$, $\rho = 4$, $B = 4000$.} \label{fig_12}
	\end{figure}

	In addition to the MNIST dataset, we conducted experiments with the CIFAR-10 dataset. For CIFAR-10 classification, we employed a convolutional 
	neural network (CNN). Since CNN does not adhere to the assumptions that allow us to determine the required number of iterations, we set the 
	number of iterations to $100$ for training CNN. With $N=20$, $\rho = 4$, and $B = 4000$, we present the results for the loss function using TDMA 
	and RA, and the completion time in Figs. \ref{fig_10}, \ref{fig_11}, and \ref{fig_12}, respectively. We consider the batch gap size $\Delta$ in the set 
	$\left\lbrace 0, 4, 8, 12, 16, 20 \right\rbrace$.
	
	As observed in the figures, the step-wise batch allocation in Algorithm~\ref{alg:sw_batch_allo} significantly reduces the completion time without 
	compromising the model's performance. In particular, the optimal batch gap is given by $\Delta = 4$ for TDMA, i.e., $\Delta=\rho$ in 
	Theorem~\ref{thm:min_titer_TDMA}. The minimum completion time is achievable when $\Delta = 8$ for RA.
	Note that the values of the optimal batch gap in Fig.~\ref{fig_12} are smaller than those in Fig.~\ref{fig_9}. This is because the computing  capability 
	of each node is reduced from $10$ to $4$, making a smaller size of the batch gap sufficient to cause a deviation in computing time across devices. 
	For instance, a batch gap larger than $20$ results in unnecessary idle time slots, leading to a longer completion time.

	\section{Conclusion}\label{sec:conclusion}
	In this paper, we thoroughly investigated completion time, representing the time required for federated learning over a wireless channel to achieve a 
	predefined performance for a machine learning model. In our pursuit of characterizing completion time, we derived the number of iterations 
	necessary to reach the desired performance level and analyzed the time duration for each iteration.
	
	In scenarios where the server can allocate time slots for communication with each device, we introduced a novel batch allocation algorithm. Our 
	research has conclusively demonstrated the optimality of this proposed algorithm in minimizing completion time.
	Moreover, when random access is employed, the use of identical batch sizes results in a high number of packet collisions, leading to extended 
	duration for each iteration. To address this issue and further reduce completion time, we introduced a batch allocation algorithm that leverages batch 
	gaps between devices. By allowing different batch sizes for each device, we successfully mitigated contention for channel access. As a result, the 
	step-wise batch allocation approach, incorporating suitable batch gaps while considering factors such as batch size, computing speed, and 
	transmission probability, emerges as a promising strategy for reducing completion time.
	
	Our findings underscore the importance of optimizing wireless federated learning systems from a holistic perspective that encompasses both 
	communication and computation aspects. The synergy between these two domains is crucial for achieving efficient and effective federated learning 
	over wireless networks.

	\bibliography{references}
	\bibliographystyle{IEEEtran}

\end{document}